%% file: arxiv.tex
\documentclass{article} 
\usepackage{arxiv,times}

\input{math_commands.tex}

\usepackage{hyperref}
\usepackage{url}
\usepackage[utf8]{inputenc}
\usepackage{graphicx}
\usepackage{subfig}
\usepackage{xcolor}
\usepackage{times}
\usepackage{amsmath, amsthm}
\usepackage{amsfonts,wrapfig,bm,color,dsfont,amssymb}
\usepackage{hyperref}
\usepackage{cleveref}
\usepackage{geometry}
\usepackage{indentfirst}
\usepackage{svg}
\usepackage{algorithm,algpseudocode}
\usepackage[tableposition = top]{caption}

\title{FORK: A FORward-looKing Actor for Model-Free Reinforcement Learning}

\author{Honghao Wei and Lei Ying\\
Department of Electrical Engineering and Compututer Science\\
The University of Michigan, Ann Arbor\\
\texttt{\{honghaow,leiying\}@umich.edu} 
}

\iclrfinalcopy 
\begin{document}

\maketitle

\begin{abstract}
In this paper, we propose a new type of Actor, named forward-looking Actor or FORK for short, for Actor-Critic algorithms.   FORK can be easily integrated into a model-free Actor-Critic algorithm. Our experiments on six Box2D and MuJoCo environments with continuous state and action spaces demonstrate significant performance improvement FORK can bring to the state-of-the-art algorithms. A variation of FORK can further solve BipedalWalkerHardcore in as few as four hours using a single GPU. 
\end{abstract}

\input{intro_arxiv}

\input{related}

\input{background}

\input{alg_arxiv}

\input{evaluation_arxiv}
\input{conclusion}

\nocite{*}

\input{bibfork.bbl}
\input{app_arxiv}

\end{document}

%% file: math_commands.tex
\usepackage{amsmath,amsfonts,bm}

\def\eqref#1{equation~\ref{#1}}
\def\Eqref#1{Equation~(\ref{#1})}

\def\1{\bm{1}}

\DeclareMathAlphabet{\mathsfit}{\encodingdefault}{\sfdefault}{m}{sl}
\SetMathAlphabet{\mathsfit}{bold}{\encodingdefault}{\sfdefault}{bx}{n}



%% file: intro_arxiv.tex
\section{Introduction}
Deep reinforcement learning has had tremendous successes, and sometimes even superhuman performance, in a wide range of applications including board games \citep{silver2016mastering},  video games \citep{vinyals2019grandmaster}, and robotics \citep{haarnoja2018composable}. A key to these recent successes is the use of deep neural networks as high-capacity function approximators that can harvest a large amount of data samples to approximate high-dimensional state or action value functions, which tackles one of the most challenging issues in reinforcement learning problems with very large state and action spaces. 

Many modern reinforcement learning algorithms are model-free, so they are applicable in different environments and can readily react to new and unseen states. This paper considers model-free reinforcement learning for problems with continuous state and action spaces, in particular, the Actor-Critic method, where Critic evaluates the state or action values of the Actor's policy and Actor improves the policy based on the value estimation from Critic. To draw an analogy between Actor-Critic algorithms and human decision making, consider the scenario where a high school student is deciding on which college to attend after graduation. The student, like Actor, is likely to make her/his decision based on the perceived values of the colleges, where the value of a college is based on many factors including (i) the quality of education it offers, its culture, and diversity, which can be viewed as instantaneous rewards of attending the college; and (ii) the career opportunities after finishing the college, which can be thought as the future cumulative reward. We now take this analogy one step further, in human decision making, we often not only consider the ``value'' of {\em current} state and action, but also further forecast the outcome of the current decision and the value of the {\em next} state. In the example above, a student often explicitly takes into consideration the first job she/he may have after finishing college, and the ``value'' of the first job. Since forward-looking is common in human decision making,  we are interested in understanding whether such forward-looking decision making can help Actor; in particular, whether it is useful for Actor to forecast the next state and use the value of future states to improve the policy.  To our great surprise, a relative straightforward implementation of forward-looking Actor, as an add-on to existing Actor algorithms, improves Actor's performance by a large margin.  

Our new Actor, named FOrward-looKing Actor or FORK for short, mimics human decision making where we think multi-step ahead. In particular, FORK includes a neural network that forecasts the next state given the current state and current action, called  {\em system network}; and a neural network that forecasts the reward given a (state, action) pair, called {\em reward network}. With the system network and reward network, FORK can forecast the next state and consider the value of the next state when improving the policy. For example, consider the Deep Deterministic Policy Gradient (DDPG) \citep{lillicrap2015continuous}, which updates the parameters of Actor as follows:
\begin{eqnarray*}
\phi \leftarrow \phi+\beta  \nabla_\phi Q_\psi(s_t,A_\phi(s_t)),
\end{eqnarray*} where $s_t$ is the state at time $t,$ $\phi$ are Actor's parameters, $\beta$ is the learning rate,  $Q_\psi(s,a)$ is the Critic network, and $A_\phi(s)$ is the Actor network. With DDPG-FORK, the parameters can be updated as follows: 
\begin{align}
\phi \leftarrow &\phi+\beta  \left(\nabla_\phi Q_\psi(s_t,A_\phi(s_t)) + \nabla_\phi R_\eta(s_t,A_\phi(s_t)) +\gamma \nabla_\phi R_\eta(\tilde{s}_{t+1},A_\phi(\tilde{s}_{t+1}))+\right.\nonumber\\
&\quad \quad \left.\gamma^2 \nabla_\phi Q_\psi(\tilde{s}_{t+2},A_\phi(\tilde{s}_{t+2}))\right),\label{flactor}
\end{align} where $R_\eta$ is the reward network, and $\tilde{s}_{t+1}$ and $\tilde{s}_{t+2}$ are the future states forecast by the system network $F_\theta.$ 

We will see that FORK can be easily incorporated into most deep Actor-Critic algorithms, by adding two additional neural networks (the system network and the reward network), and by adding extra terms to the loss function when training Actor, e.g. adding term $$R_\eta(s_t,A_\phi(s_t)) +\gamma R_\eta(\tilde{s}_{t+1},A_\phi(\tilde{s}_{t+1}))+\gamma^2  Q_\psi(\tilde{s}_{t+2},A_\phi(\tilde{s}_{t+2}))$$ for each sampled state $s_t$ to implement (\ref{flactor}).  

We remark that \Eqref{flactor} is just one example of FORK, FORK can have different implementations (a detailed discussion can be found in Section \ref{sec:alg}). We further remark that learning the system model is not a new idea and has a long history in reinforcement learning, called model-based reinforcement learning (some state-of-the-art model-based reinforcement learning algorithms and the benchmark can be found in \citep{WanBaoCla_19}). Model-based reinforcement learning uses the model in a sophisticated way, often based on deterministic or stochastic optimal control theory to optimize the policy based on the model. FORK only uses the system network as a blackbox to forecast future states, and does not use it as a mathematical model for optimizing control actions. With this key distinction, any model-free Actor-Critic algorithm with FORK remains to be model-free.  

In our experiments, we added FORK to two state-of-the-art model-free algorithms, according to recent benchmark studies \cite{duan2016benchmarking,WanBaoCla_19}: TD3 \cite{fujimoto2018addressing} (for deterministic policies) and SAC \cite{HaaZhoAur_18} (for stochastic policies).  The evaluations on six challenging environments with continuous state space and action space show significant improvement when adding FORK. In particular, SAC-FORK performs the best among five of the six environments, and TD3-FORK performs the best for the rest one. For Ant-v3, it improves the average cumulative reward by more than $50\%$ than TD3, and achieves TD3's best performance using only $35\%$ of training samples. BipedalWalker-v3 is considered ``solved'' when the agent obtains an average cumulative reward of at least 300\footnote{\url{https://github.com/openai/gym/blob/master/gym/envs/box2d/bipedal_walker.py}}. TD3-FORK only needs 0.47 million training steps to solve the problem, half of that under TD3. Furthermore, a variation of TD3-FORK solves  BipedalWalkerHardcore, a well known difficult environment,  with as few as four hours using a single GPU. The source code of our FORK implementation\footnote{\url{https://github.com/honghaow/FORK}} is available online.

%% file: related.tex
\subsection{Related Work}
\label{sec:related}

The idea of using learned models in reinforcement learning is not new, and actually has a long history in reinforcement learning. At a high level, FORK shares a similar spirit as model-based reinforcement learning and rollout.  However, in terms of implementation, FORK is very different and much simpler. Rollout in general requires the Monte-Carlo method \citep{silver2017mastering} to simulate a finite number of future states  from the current state and then combines that with value function approximations to decide the action to take at the current time. FORK does not require any high-fidelity simulation. The key distinction between FORK and model-based reinforcement learning is that model-based reinforcement learning uses the learned model in a sophisticated manner. For example, in SVG \cite{HeeWaySil_15}, the learned system model is integrated as a part of the calculation of the value gradient,  in \cite{gu2016continuous}, refitted local linear model and rollout are used to derive linear-Gaussian controller, and \cite{bansal2017mbmf} uses a learned dynamical model to compute the trajectory distribution of a given policy and consequently estimates the corresponding cost using a Bayesian optimization-based policy search. More model-based reinforcement learning algorithms and related benchmarking can be found in \cite{WanBaoCla_19}. With knowing the closed from of underlying reward functions, \cite{KurClaDua_18,LuoXuLi_18} apply the learned model to generate imagined data, and then use these additional traces in the policy optimization step. FORK, on the other hand, only uses the system network to predict future states, and does not use the system model beyond that. Other related work that accelerates reinforcement learning algorithms includes: acceleration through exploration strategies \cite{gupta2018meta}, optimizers \cite{duan2016rl}, and intrinsic reward \cite{zheng2018learning}, just to name a few. These approaches are complementary to ours. FORK can be added to further accelerate learning.

%% file: background.tex
\section{Background}

Reinforcement Learning algorithms aim at learning policies that maximize the cumulative reward by interacting with the environment. We consider a standard reinforcement learning setting defined by a Markov decision process (MDP) $(\mathcal{S},\mathcal{A},p_0,p,r,\gamma),$ where $\mathcal{S}$ is a set of states, $\mathcal{A}$ is the action space, $p_0(s)$ is the probability distribution of the initial state, $p:\mathcal{S}\times\mathcal{S}\times\mathcal{A}\rightarrow[0,\infty)$ is the transition density function, which represents the distribution of the next state $s_{t+1}$ given current state $s_t$ and action $a_t$, $r:\mathcal{S}\times\mathcal{A}\rightarrow [r_{\min},r_{\max}]$ is the bounded reward function on each transition, and $\gamma \in (0,1]$ is the discount factor.  We consider a discrete-time system. At each time step $t,$ given the current $s_t\in \mathcal{S},$ the agent selects an action $a_t\in\mathcal{A}$ based on a (deterministic or stochastic) policy $\pi(a_t\vert s_t),$ which moves the environment to the next state $s_{t+1},$ and yields a reward $r_t=r(s_t,a_t)$ to the agent. We consider stationary policies in this paper under which the action is taken based on $s_t,$ and is independent of other historical information. 

Starting from time $0,$ the return of given policy $\pi$ is the discounted cumulative reward $$J_\pi(i)=\sum_{t=0}^T \gamma^{t}r(s_t,a_t), \quad\hbox{given } s_0=i.$$ $J_\pi(i)$ is also called the state-value function. Our goal is to learn a policy $\pi^*$ that maximizes this cumulative reward $$\pi^*\in\arg\max_\pi J_\pi(i) \quad \forall i.$$ We assume our policy is parameterized by parameter $\phi,$ denoted by $\pi_\phi,$ e.g. by the Actor network in Actor-Critic Algorithms.  In this case, our goal is to identify the optimal parameter $\phi^*$ that maximizes $$\phi^*\in\arg\max J_{\pi_\phi}(i).$$ 

Instead of state-value function, it is often convenient to work with action-value function, Q-function, which is defined as follows: 
$$Q_\pi(s,a)= E\left[r_\pi(s,a) + \gamma J_\pi(s')\right],$$ where $s'$ is the next state given current state $s$ and action $a.$ The optimal policy is a policy that satisfies the following Bellman equation \citep{BellmanDynamicProgramming}: $$Q_{\pi^*}(s,a)= E\left[r(s,a) + \gamma \max_{a^\prime\in\mathcal{A}}Q_{\pi^*}({s^\prime,a^\prime})\right].$$ When neural networks are used to approximate action-value functions, we denote the action-value function by $Q_\psi(s,a),$ where $\psi$ is the parameters of the neural network.

%% file: alg_arxiv.tex
 \section{FORK --- Forward-Looking Actor}
 \label{sec:alg}
This paper focuses on Actor-Critic algorithms, where Critic estimates the state or action value functions  of the current policy, and Actor improves the policy based on the value functions. 
We propose a new type of Actor, FORK. More precisely, a new training algorithm that improves the policy by considering not only the action-value of the current state (or states of the current mini-batch), but also future states and actions forecast using a learned system model and a learned reward model. This forward-looking Actor is illustrated in Figure \ref{fig:fl-actor}. In FORK, we introduce two additional neural networks:

\begin{wrapfigure}{r}{0.5\textwidth}
    \centering
    \includegraphics[width=3in]{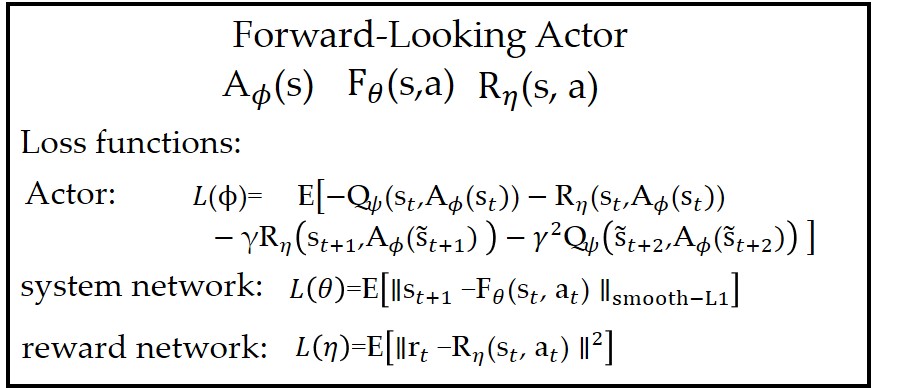}
    \caption{FORK includes three neural networks, the policy network $A_\phi,$ the system model $F_\theta,$ and the reward model $R_\eta$.}
    \label{fig:fl-actor}
\end{wrapfigure}

\noindent{\bf The system network $F_\theta$.} The network is used to predict the next state of the environment, i.e., given current state $s_t$ and action $a_t,$ it predicts the next state $\tilde{s}_{t+1}= F_\theta(s_t,a_t).$ With experiences $(s_t,a_t,s_{t+1}),$ training the system network is a supervised learning problem. The neural network can be trained using mini-batch from replay-buffer and smooth-L1 loss 
    $L(\theta)= \|s_{t+1}-F_\theta(s_t, a_t)\|_{\hbox{\scriptsize smooth L1}}.$
    
\noindent {\bf The reward network $R_\eta.$} This network predicts the reward given current state $s_t$ and action $a_t,$ i.e. $\tilde{r}_t=R_\eta(s_t, a_t).$ The network can be trained from experience $(s_t,a_t,r_t),$ with MSE loss 
    $L(\eta)=\|r_{t}-R_\eta(s_t,a_t)\|^2.$

{\bf FORK.} With the system network and the reward network, the agent can forecast the next state, the next next states and so on. Actor can then use the forecast to improve the policy. For example, we consider the following loss function 
\begin{align}
    L(\phi)=E\left[-Q_\psi(s_t, A_\phi(s_t))-R_\eta(s_t, A_\phi(s_t))-\gamma R_\eta(\tilde{s}_{t+1}, A_\phi(\tilde{s}_{t+1}))-\gamma^2 Q_\psi\left(\tilde{s}_{t+2}, A_\phi(\tilde{s}_{t+2})\right)\right].
    \label{loss:actor}
\end{align}
In the loss function above, $s_t$ are from data samples (e.g. replay buffer), $\tilde{s}_{t+1}$ and $\tilde{s}_{t+2}$ are calculated from the system network as shown below: 
\begin{align}
\tilde{s}_{t+1}=F_\theta(s_t,A_\phi(s_t))\quad\hbox{ and }\quad \tilde{s}_{t+2}=F_\theta(\tilde{s}_{t+1},A_\phi(\tilde{s}_{t+1})). \label{eq:future-states}
\end{align}
Note that when training Actor $A_\phi$ with loss function $L(\phi),$ all other parameters in $L(\phi)$ are regarded as constants except $\phi.$ 

The action-function $Q,$ without function approximation, under current policy $A_\phi$ satisfies 
\begin{align*}
        Q(s_t, A_\phi(s))
        =E\left[r(s_t, A_\phi(s_t))+\gamma r({s}_{t+1}, A_\phi({s}_{t+1}))+\gamma^2 Q\left({s}_{t+2}, A_\phi({s}_{t+2})\right)\right],
\end{align*} where $r,$ $s_{t+1}$ and $s_{t+2}$ are the actual rewards and states under the current policy, not estimated values. Therefore, the loss function $L(\phi)$ can be viewed as the average of two estimators.

Given action values from Critic and with a mini-batch of size $N,$ FORK updates its parameters as follows:
$$\phi \leftarrow \phi - \beta_t \triangledown_\phi L(\phi),$$ where $\beta_t$ is the learning rate and 
\begin{align*}
                \triangledown_\phi L(\phi) =
                &\frac{1}{N}\sum_{i=1}^N \left( \nabla_{a}Q_{\psi}(s_i,a)|_{a=A_\phi(s_i)} \nabla_\phi A_\phi(s_i)+ \nabla_{a} R_\eta(s_i,a) |_{a=A_\phi(s_i)} \nabla_\phi A_\phi(s_i) \right.\\
                &\left.+\gamma \nabla_{a} R_\eta(\tilde{s}'_i,a)|_{a=A_\phi(\tilde{s}'_i)} \nabla_\phi A_\phi(\tilde{s}'_i) + \gamma^2 \nabla_{a} Q_{\psi}(\tilde{s}_{i}'',a) |_{a=A_\phi(\tilde{s}''_i)} \nabla_\phi A_\phi(\tilde{s}''_i) \right),
\end{align*} where $\tilde{s}_i'$ and $\tilde{s}_i''$ are the next state and the next next state estimated from the system network.

\begin{wrapfigure}{r}{0.4\textwidth}
    \centering
    \includegraphics[scale=0.3]{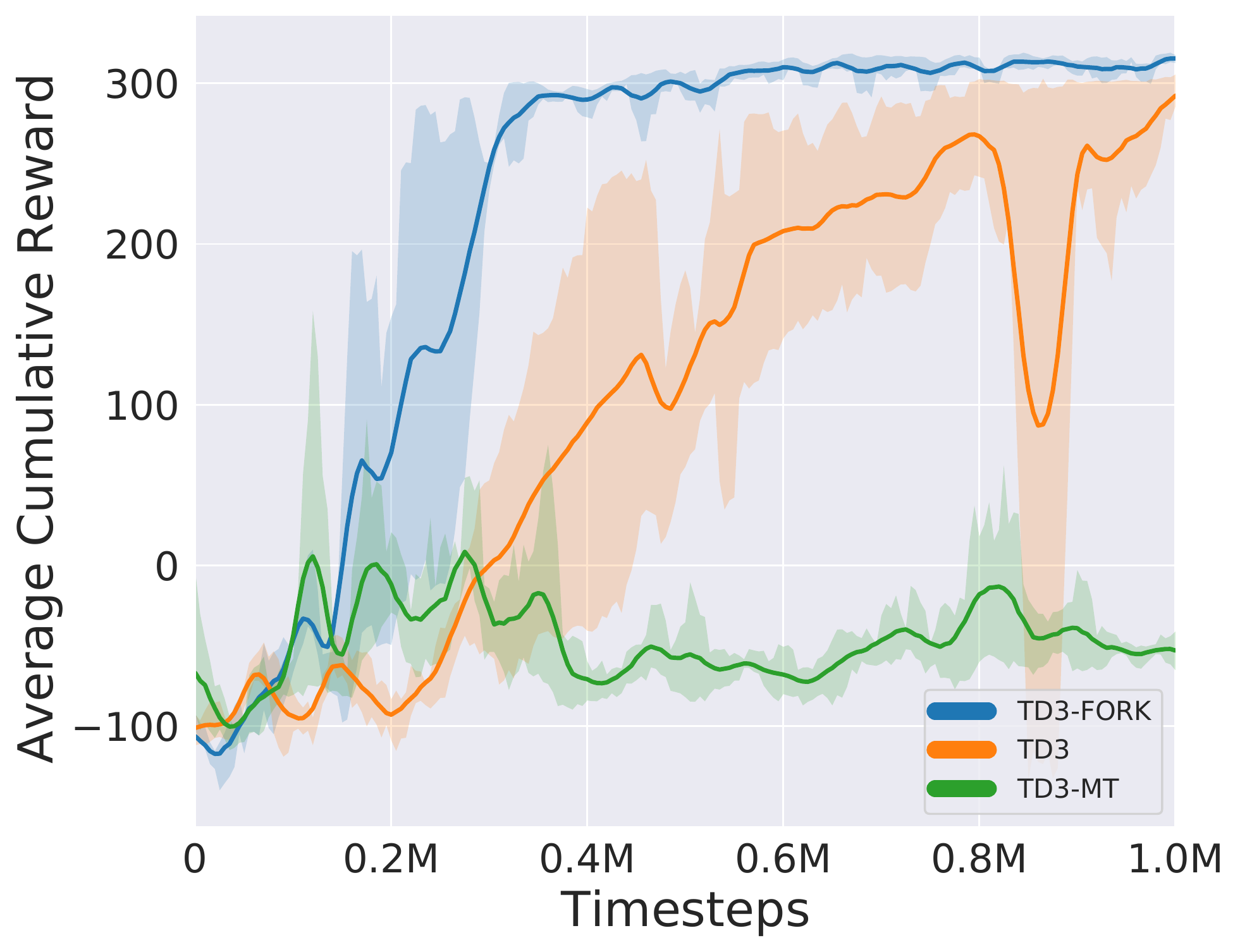}
    \caption{TD3-FORK versus TD3-MT}
    \label{fig:multiple_transitions}
    \vspace{-0.2in}
\end{wrapfigure}

We note that it is important to use the system network to generate future states as in \Eqref{eq:future-states} because they mimic the states under the current policy. If we would sample a sequence of consecutive states from the replay buffer, then the sequence is from the old policy, which  does not help the learning. Figure \ref{fig:multiple_transitions} compares average performance of TD3-FORK, TD3, and TD3-MT, which samples a sequence of three consecutive states, on the BipedalWalker-v3 environment over 2 instances. We can clearly see that simply using consecutive states from experiences does not help improve learning. In fact, it significantly hurts the learning.

\noindent{\bf Modified Reward Model:} We found from our experiments that the reward network can more accurately predict reward $r_t$ when including the next state $s_{t+1}$ as input into the reward network (an example can found in Appendix \ref{appsec:revised-reward}). Therefore, we use a modified reward network $R_\eta(s_t, a_t, s_{t+1})$ in FORK. 

\noindent{\bf Adaptive Weight:} Loss function $L(\phi)$  in our algorithm uses the system network and the reward network to boost learning. In our experiments, we found that the forecasting can significantly improve the performance, except at the end of learning. Since the system and reward networks are not perfect, the errors in prediction can introduce errors/noises. To overcome this issue, we found it is helpful to use an adaptive weight $w$ so that FORK accelerates learning at the beginning but its weight decreases gradually as it gets close to the learning goal. A comparison between fixed weights and adaptive weights can be found in Appendix \ref{appsec:fixed-adaptive}. We use a simple adaptive weight $w = \left(\frac{\bar{r}}{r_0}\right)_0^1w_0,$ where $\bar{r}$ is the moving average of cumulative reward (per episode),  and $r_0$ is a predefined goal, $w_0$ is the initial weight, and $(a)_0^1=a$ if $0\leq a\leq 1,$ $=0$ if $a<0$ and $=1$ if $a>1.$ The loss function with adaptive weight becomes 
\begin{align}
    L(\phi)=E\left[-Q_\psi(s_t, A_\phi(s_t))-wR_\eta(s_t, A_\phi(s_t))-w\gamma R_\eta(\tilde{s}_{t+1}, A_\phi(\tilde{s}_{t+1}))-w\gamma^2 Q_\psi\left(\tilde{s}_{t+2}, A_\phi(\tilde{s}_{t+2})\right)\right].
    \label{loss:actor-aw}
\end{align}
Furthermore, we set a threshold and let $w=0$ if the loss of the system network is larger than the threshold. This is to avoid using FORK when the system and reward networks are very noisy. We note that in our experiments, the thresholds were chosen such that $w=0$ for around $20,000$ steps at the beginning of each instance, which includes the first 10,000 random exploration steps.

\noindent{\bf Different Implementations of FORK:} It is easy to see FORK can be implemented in different forms. For example, instead of two-step ahead, we can use one-step ahead as follows: 
\begin{align}
    L(\phi)=E\left[-Q_\psi(s_t, A_\phi(s_t))-w R_\eta(s_t, A_\phi(s_t))-w \gamma Q_\psi\left(\tilde{s}_{t+1}, A_\phi(\tilde{s}_{t+1})\right)\right], \label{loss:fork-s}
\end{align} or only use future action values: 
\begin{align}
    L(\phi)=E\left[-Q_\psi(s_t, A_\phi(s_t))- w\left( Q_\psi\left(\tilde{s}_{t+1}, A_\phi(\tilde{s}_{t+1})\right)-w^\prime Q_\psi\left(\tilde{s}_{t+2}, A_\phi(\tilde{s}_{t+2})\right)\right)\right].\label{loss:fork-dq}
\end{align} We compared these two versions with FORK. The performance comparison can be found in Appendix \ref{appsec:forks}.

%% file: evaluation_arxiv.tex
\section{Experiments}
\label{sec:eval}
In this section, we evaluate FORK as an add-on to existing algorithms. We name an algorithm with FORK as algorithm-FORK, e.g. TD3-FORK or SAC-FORK. As an example, a detailed description of TD3-FORK can be found in Appendix \ref{appsec:td3fork}. 
We focused on two  algorithms: TD3 \citep{fujimoto2018addressing} and SAC \citep{HaaZhoAur_18} because they were found to have the best performance among model-free reinforcement learning algorithms in recent  benchmarking studies  \citep{duan2016benchmarking,WanBaoCla_19}.    We compared the performance of TD3-FORK and SAC-FORK with TD3, SAC and  DDPG \citep{lillicrap2015continuous}.  

\subsection{Box2D and MuJoCo Environments}

\begin{figure}[tb]
    \centering
     \subfloat[BipedalWalker-v3]{{\includegraphics[width=2.5cm,height=2.5cm]{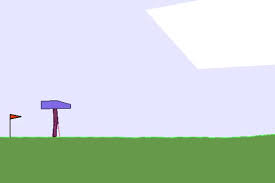} }}
    \subfloat[Ant-v3]{{\includegraphics[width=2.5cm,height=2.5cm]{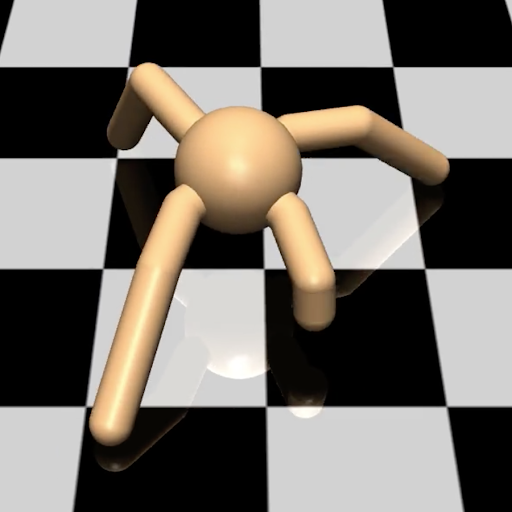}}}
    \subfloat[Hopper-v3]{{\includegraphics[width=2.5cm,height=2.5cm]{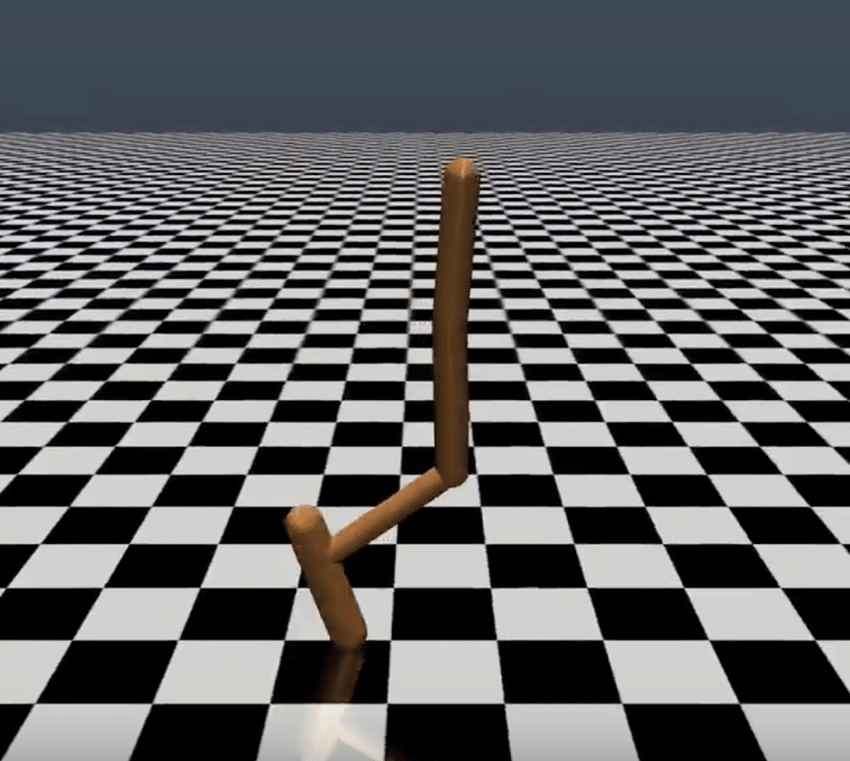}}}
    \subfloat[HalfCheetah-v3]{{\includegraphics[width=2.5cm,height=2.5cm]{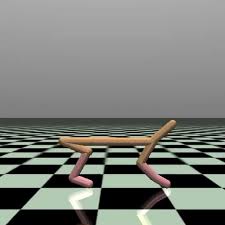} }}
     \subfloat[Humanoid-v3]{{\includegraphics[width=2.5cm,height=2.5cm]{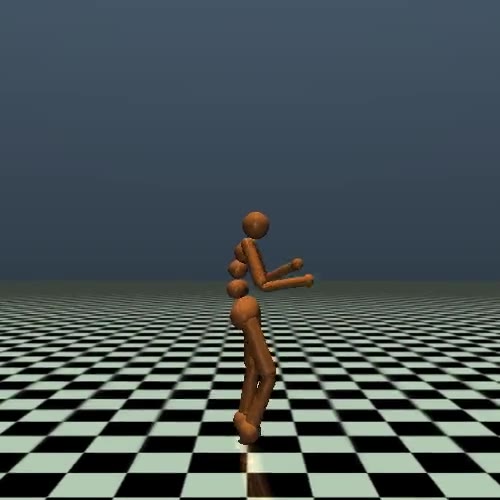} }}
     \subfloat[Walker2d-v3]{{\includegraphics[width=2.5cm,height=2.5cm]{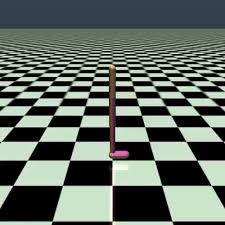} }}
    \caption{The six environment used in our experiments}
    \label{fig:mujoco-env}
\end{figure}

We selected six environments: BipedalWalker-v3 from Box2D \citep{Catbox}, Ant-v3, Hooper-v3, HalfCheetah-v3, Humanoid-v3 and Walker2d-v3 from MuJoCo \citep{todorov2012mujoco} as shown in  Figure \ref{fig:mujoco-env}.  All these environments have continuous state spaces and action spaces.

\subsection{Implementation Details}

{\bf Terminology.} {\bf step (or time)}: one operation, e.g. interaction with the environment once; {\bf episode:} a single-run of the environment from the beginning to the end, consisting of many steps; and {\bf instance:} the entire training consisting of multiple episodes.  

{\bf Hyperparameters.} Because FORK is an add-on, for TD3, we used the authors' implementation (\url{https://github.com/sfujim/TD3}); for SAC, we used a PyTorch version (\url{https://github.com/vitchyr/rlkit}) recommended by the authors  {\em without any change except adding FORK}. The hyperparameters of both TD3 and SAC are summarized in Table \ref{tab:hyper} in Appendix \ref{appsec:hyper}, and the hyperparameters related to FORK are summarized in Table \ref{tab:hyper-2} in the same appendix. 

We can see TD3-FORK does not require much  hyperparameter tuning. The system network and reward network used in the environments are the same except for the Humanoid-v3 for which we use larger system and reward networks because the dimension of the system is higher than other systems.  The base weight $w_0$ is the same for all environments, the base rewards are the typical cumulative rewards under TD3 after a successful training, and the system thresholds are the typical estimation errors after about 20,000 steps. 

SAC-FORK requires slightly more hyperparameter tuning. The base weights were chosen to be smaller values, the base rewards are the typical cumulative rewards under SAC, and the system thresholds are the same as those under TD3-FORK. 

{\bf Initial Exploration.} For each task and each algorithm, we use a random policy for exploration for the first 10,000 steps. Each step is one interaction with the environment.  

{\bf Duration of Experiments.} For each environment and each algorithm, we ran five different instances with different random seeds. Each instance is run for 1 million time steps. 

\subsection{Results}
\begin{figure*}[tb]
    \centering
    \subfloat[BipedalWalker-v3]{{\includegraphics[scale=0.25]{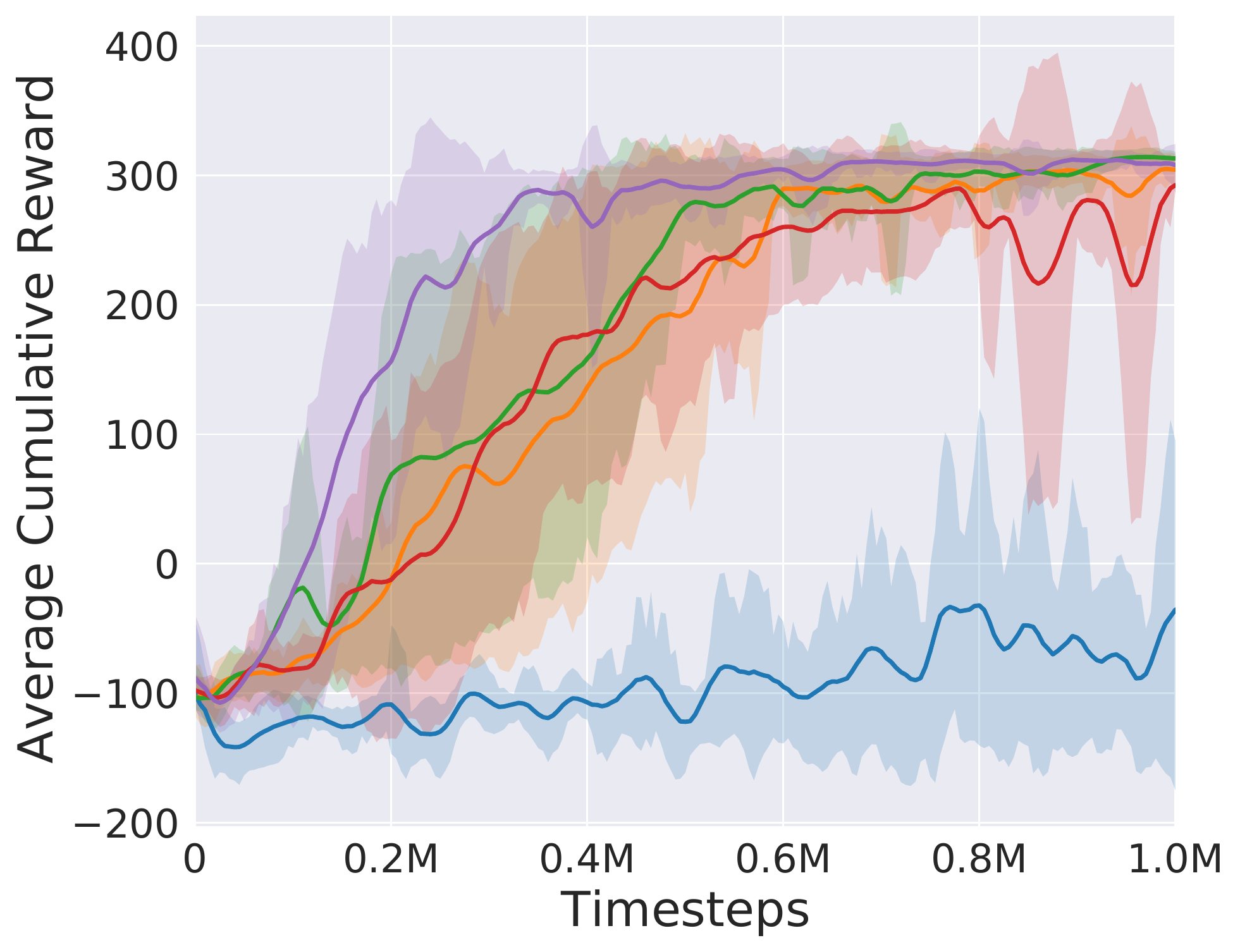} }}
    \subfloat[Ant-v3]{{\includegraphics[scale=0.25]{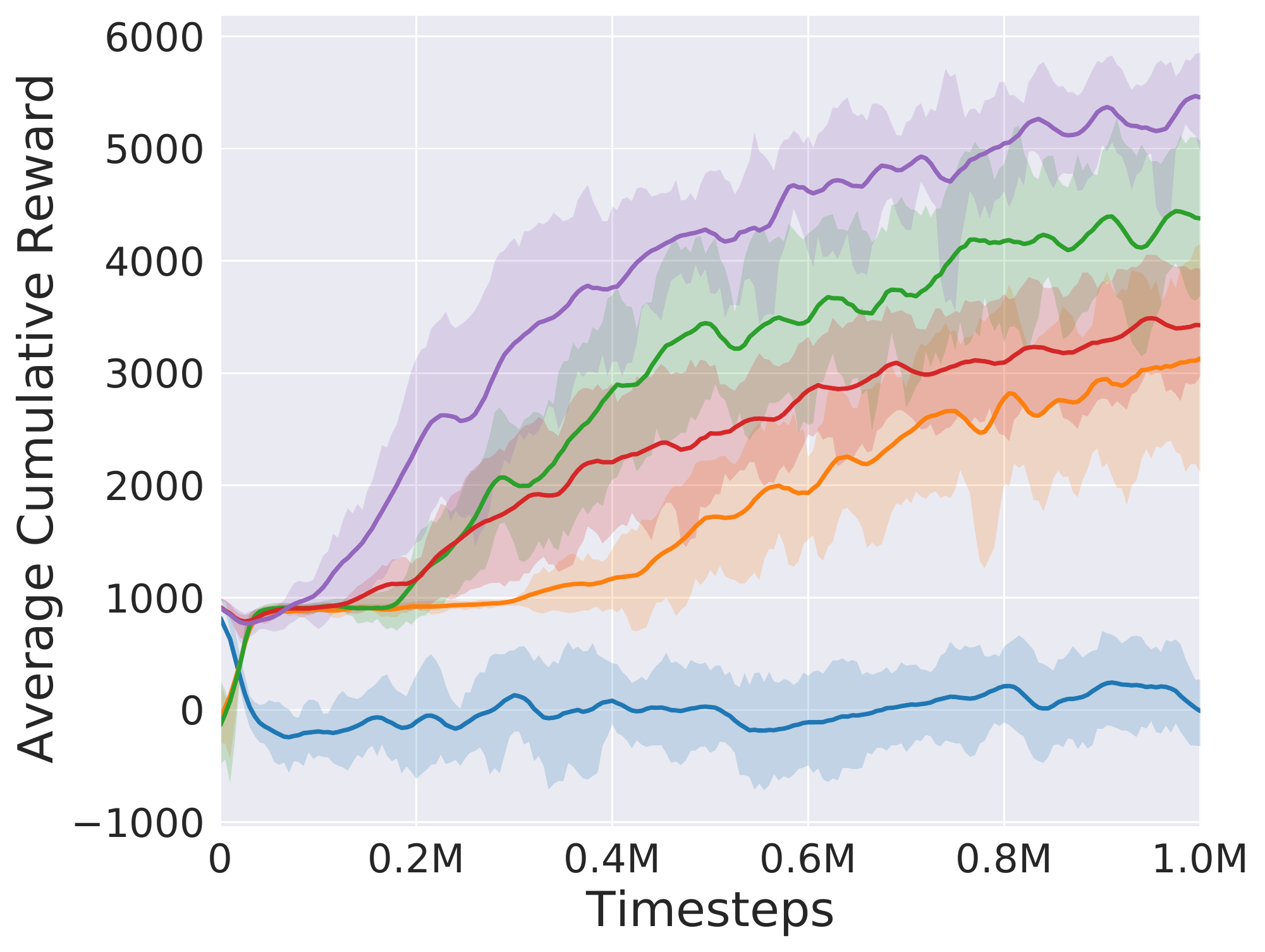} }}
    \subfloat[Hopper-v3]{{\includegraphics[scale=0.25]{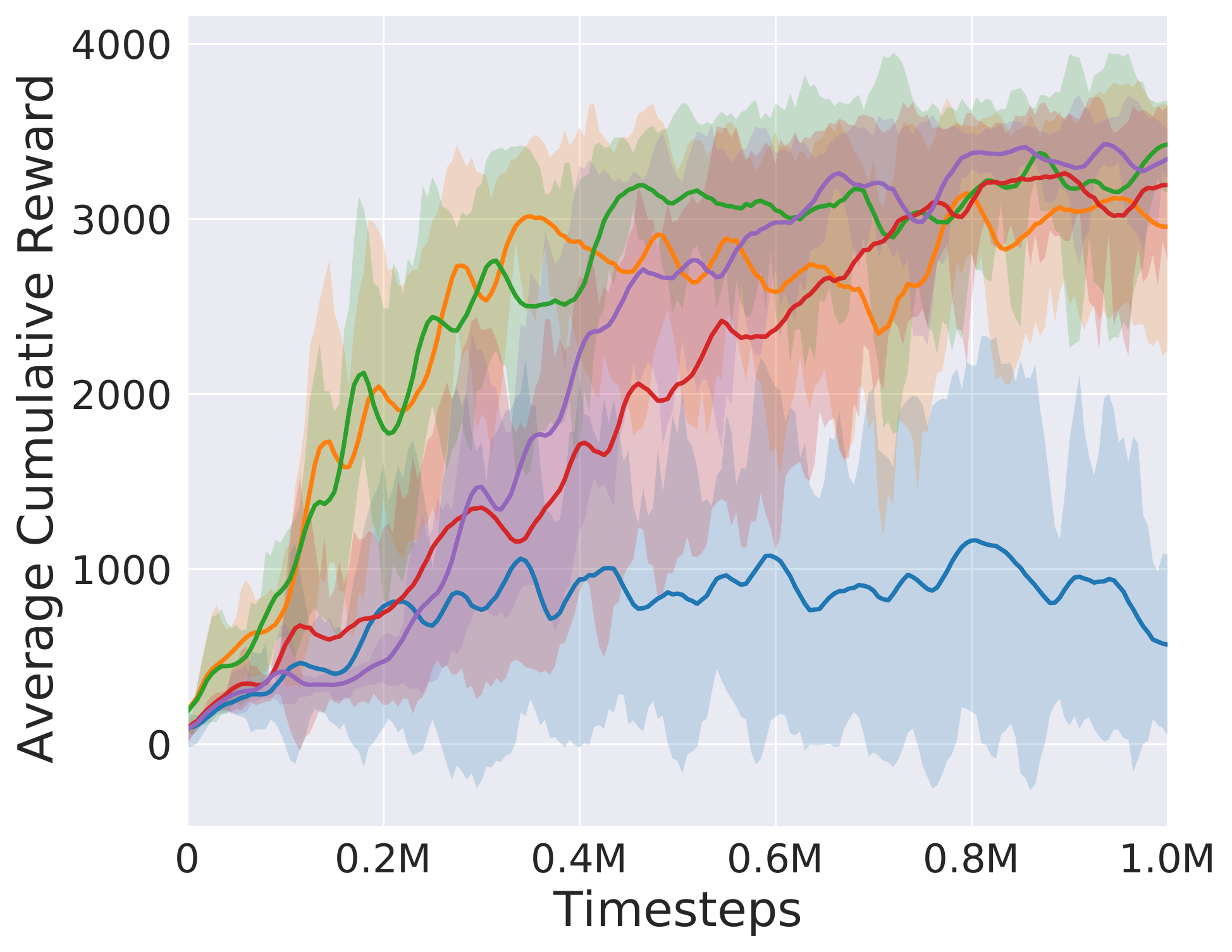} }} \quad
    \subfloat[HalfCheetah-v3]{{\includegraphics[scale=0.25]{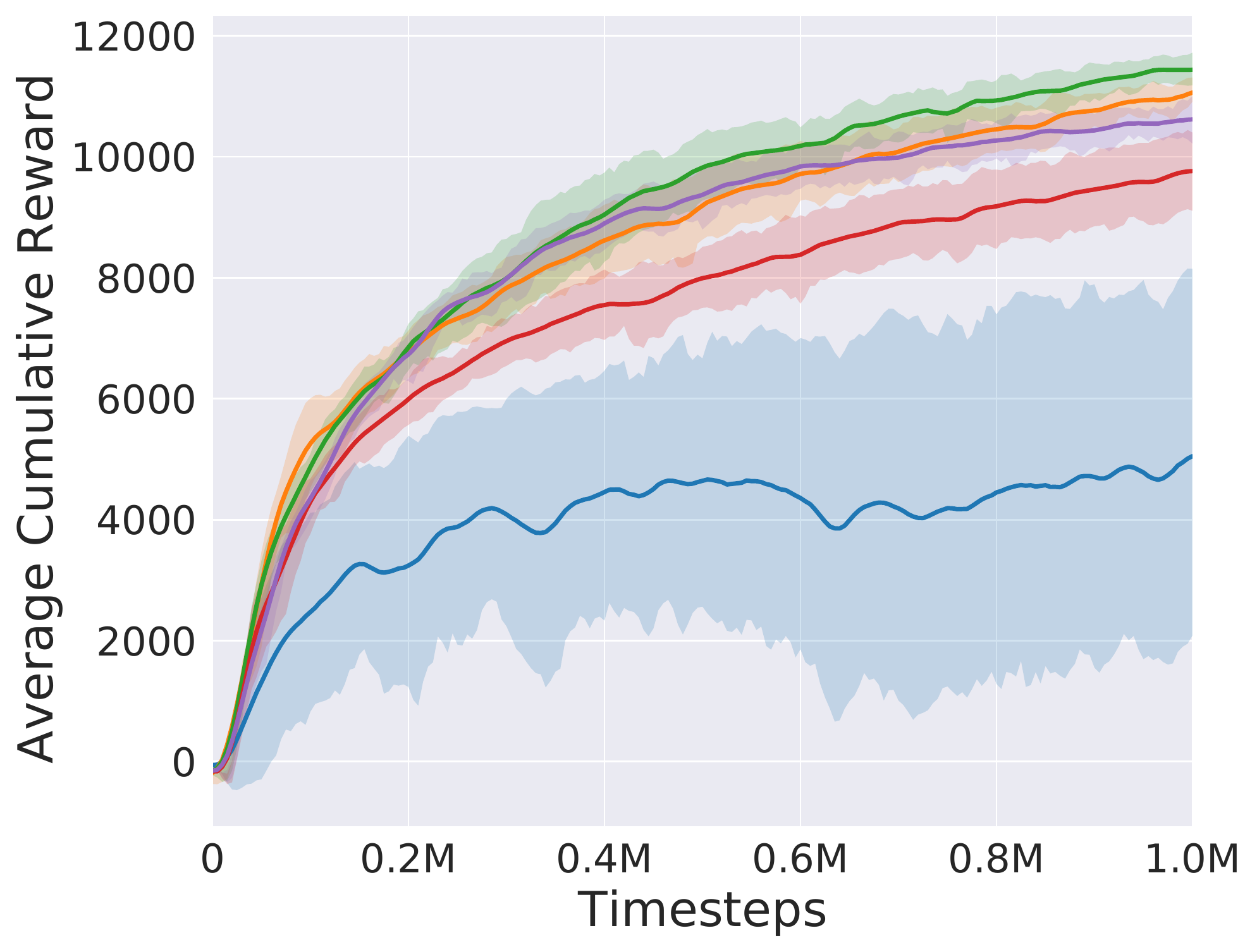}}}
     \subfloat[Humanoid-v3]{{\includegraphics[scale=0.25]{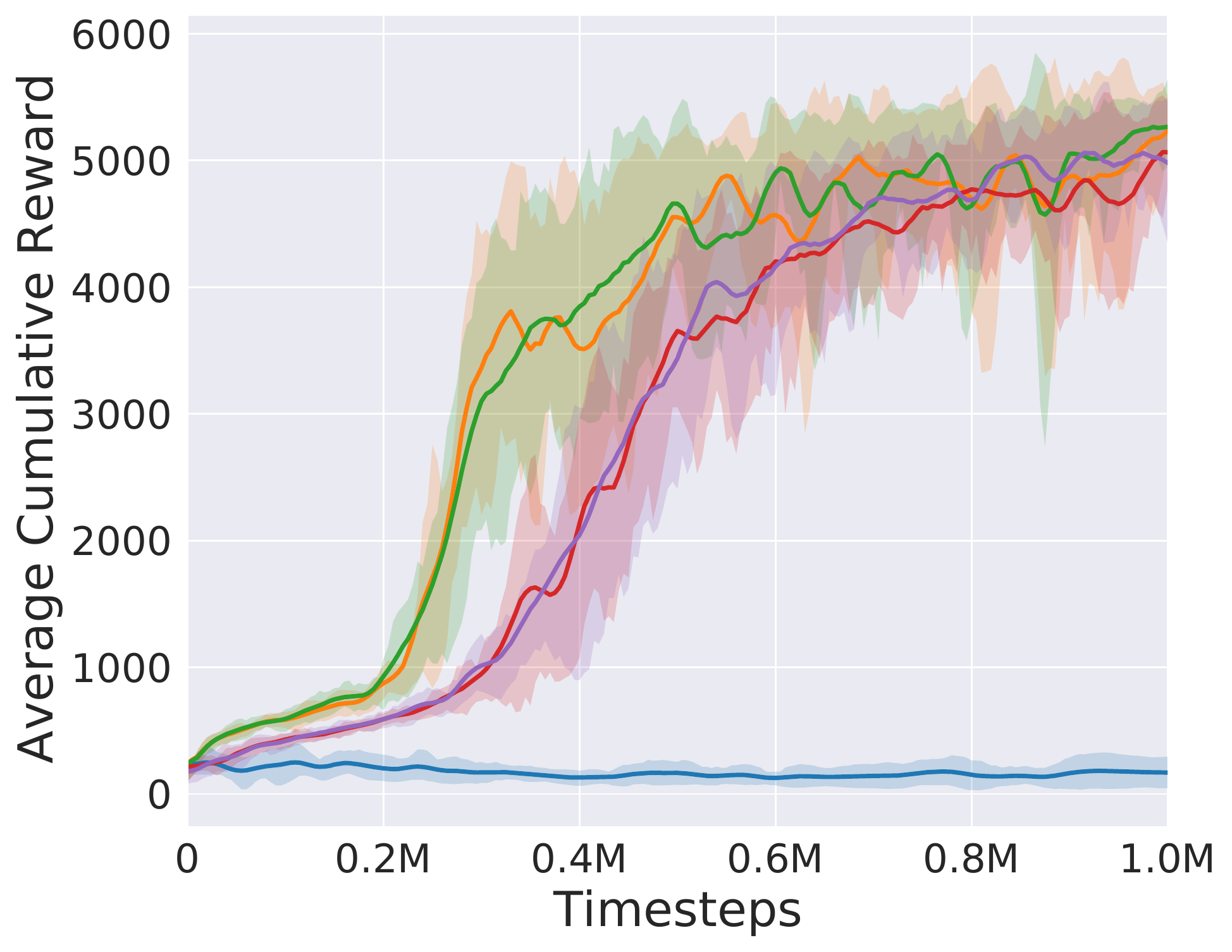} }}
    \subfloat[Walker2d-v3]{{\includegraphics[scale=0.25]{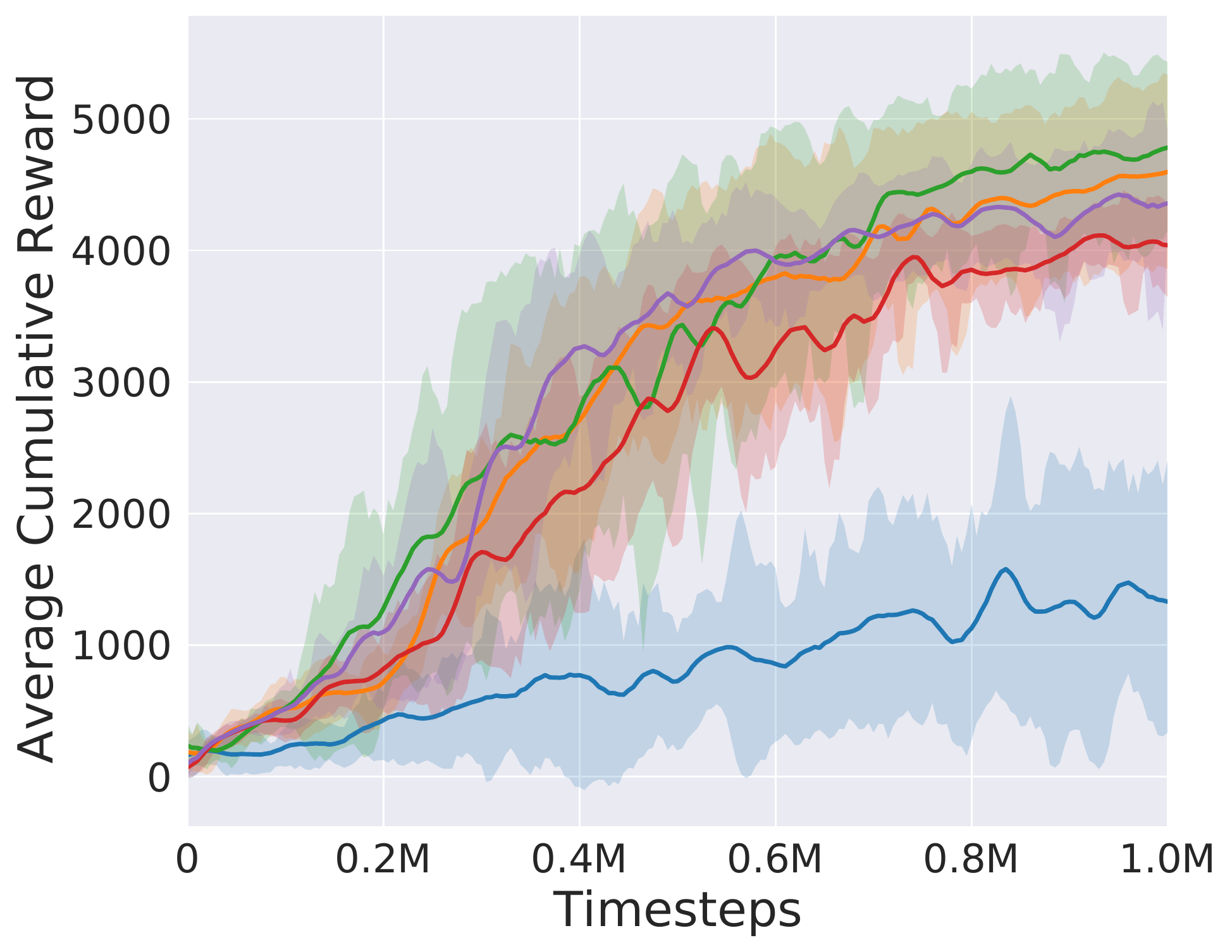} }}\quad 
    \subfloat{{\includegraphics[scale=0.5]{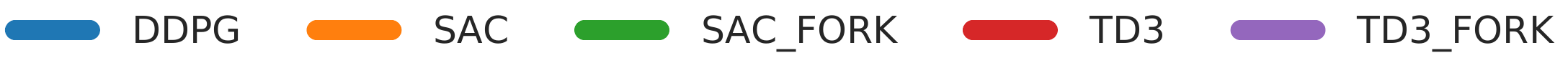} }}
    \caption{Learning curves of six environments. Curves were smoothed uniformly for visual clarity.}
    \label{fig:mujoco}
     \vspace{-0.2in}
\end{figure*}
Figure \ref{fig:mujoco} shows the average cumulative rewards, where we evaluated the policies every 5,000 steps without exploration noise during training process. Each evaluation was averaged over 10 episodes. We train five different instances for each algorithm with same random seeds. The solid curves shows the average cumulative rewards (per episode), and the shaded region represents the standard deviations. 

The best average cumulative rewards are summarized in Table \ref{tab:max}. We can see that SAC-FORK outperform all other algorithms in five of the six environments and TD3-FORK achieves the best cumulative rewards in the remaining one. Both algorithms have improvements compared with the original ones without FORK. For Ant-v3, TD3-FORK improves the best average cumulative reward by more than 50\% (5731.85 (TD3-FORK) versus 3678.75 (TD3)). 

\begin{table*}[t]
    \centering
    \caption{The best average cumulative rewards of the algorithms. The best value for each environment is highlighted in bold text.}
    \begin{tabular}{lccccc}
     \hline
        {\bf Environment} &  {\bf TD3-FORK} & {\bf TD3} & {\bf DDPG} & {\bf SAC} & {\bf SAC-FORK} \\
    \hline
    BipedalWalker-v3 & ${ 317.73}$   & $307.97$ &   $139.29$  & $313.49$ & ${\bf 319.47}$\\
    \hline
    Ant-v3& ${\bf 5731.85}$ & $3678.75$   &  $995.59$  &  $3600.24$ &  $5033.09$  \\
    \hline
    Hopper-v3 & ${3539.95} $ & $3526.67 $ & $1738.05$ &  $3504.41$ & ${\bf 3568.20}$ \\
    \hline
    HalfCheetah-v3 & ${10875.32} $   & $9899.12$  & $5975.80 $ & $11230.00$ & ${\bf 11604.60}$\\
    \hline
     Humanoid-v3 & ${5439.31} $  & $5394.02$    & $465.10$ & $5484.04 $ & ${\bf 5509.89}$ \\
    \hline
    Walker2d-v3 &  ${ 4619.64 }$   &    $4319.09 $ &  $ 2505.23  $ & $ 4763.36$ & $  {\bf 5093.01}$  \\
    \hline
    \end{tabular}
    \label{tab:max}
\end{table*}

We also studied the improvement in terms of sample complexity. In Table \ref{tab:data-ef}, we summarized the number of Actor training required  under TD3-FORK (SAC-FORK) to achieve the best average cumulative reward under TD3 (SAC). For example, for BipedalWalker-v3, TD3 achieved the best average cumulative reward with 0.985 million steps; and TD3-FORK achieved the same value with only 0.470 million steps, reducing the required samples by more than 50\%.

 \begin{table}[tb]
    \centering
    \caption{Sample complexity (million). The number of training steps needed for TD3-FORK (SAC-FORK) to achieve the same best average cumulative reward under TD3 (SAC). The numbers under TD3 (SAC) are the time steps at which the TD3 (SAC) achieved the best average cumulative rewards.}
    \begin{tabular}{lcccc}
     \hline
    {\bf Environment} &  {\bf TD3-FORK} & {\bf TD3}&{\bf SAC-FORK} & {\bf SAC }\\
    \hline
    BipedalWalker-v3 & 0.470 & 0.985 & 0.885 & 0.960\\
    \hline
    Ant-v3 & 0.345 & 0.950 & 0.440 & 0.950 \\
    \hline
    Hopper-v3 &  0.810 & 0.915 & 0.560 & 0.810\\
    \hline
     HalfCheetah-v3 & 0.590 & 0.990 & 0.850 & 0.995\\
     \hline
     Humanoid-v3 & 0.860 & 0.995 & 1.000& 1.000 \\
     \hline
     Walker2d-v3 & 0.720 & 1.000 & 0.785 & 0.975\\
     \hline
    \end{tabular}
    \label{tab:data-ef}
\end{table}

In summary, FORK improves the performance of both TD3 and SAC after being included as an add-on. 
The improvement is more significant when adding to TD3 than adding to SAC. FORK improves both SAC and TD3 in all six environments. More statistics about this set of experiments can be found in  Appendix \ref{appsec:more}.

\subsection{Comparison with Model-Based Reinforcement Learning Algorithms}

We also compared the performance and time complexity of FORK with the following model-based algorithms:
\begin{itemize}
	\item Model-Ensemble Trust-Region Policy Optimization (ME-TRPO) \cite{KurClaDua_18}: ME-TRPO uses an ensemble of neural networks to model the system, which can address the model-bias. Trust-Region Policy Optimization (TRPO) \cite{SchLevPie_15} is used in the policy improvement step on the imagined data generated by the learned models.
	\item Stochastic Lower Bound Optimization (SLBO) \cite{LuoXuLi_18}: SLBO is a variant of ME-TRPO which uses a multi-step L2-norm loss to train the system dynamics. It optimizes the model and policy network alternatively during each iteration.
	\item MB-TRPO \cite{LuoXuLi_18}: A algorithm that is similar to SLBO but optimizes the policy and model together during one iteration. 
\end{itemize}

We evaluated all the algorithms mentioned above along with SAC and SAC-FORK on two environments Walker2d and HalfCheetah (different from Walker2d-v3 and HalfCheetah-v3) which have been widely used for benchmarking model-based algorithms \cite{WanBaoCla_19}. In \cite{WanBaoCla_19}, the authors benchmark model-based and model-free algorithms with unified settings to obtain a fair comparison.
 
More details could be found in \cite{WanBaoCla_19}.

\begin{figure}[!htb]
	\centering
	\subfloat[HalfCheetah]{{\includegraphics[scale=0.35]{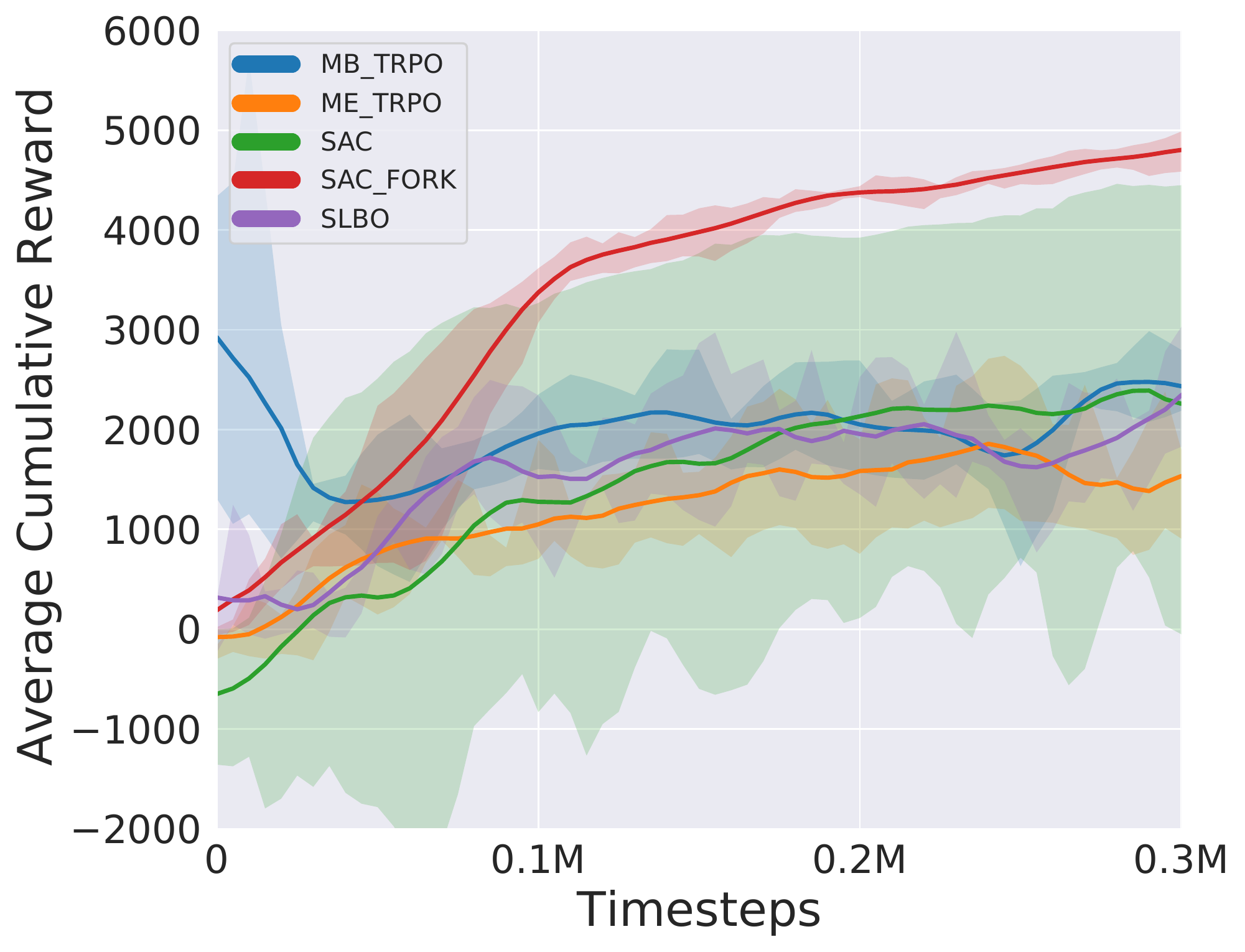} }}
	\subfloat[Walker2d]{{\includegraphics[scale=0.35]{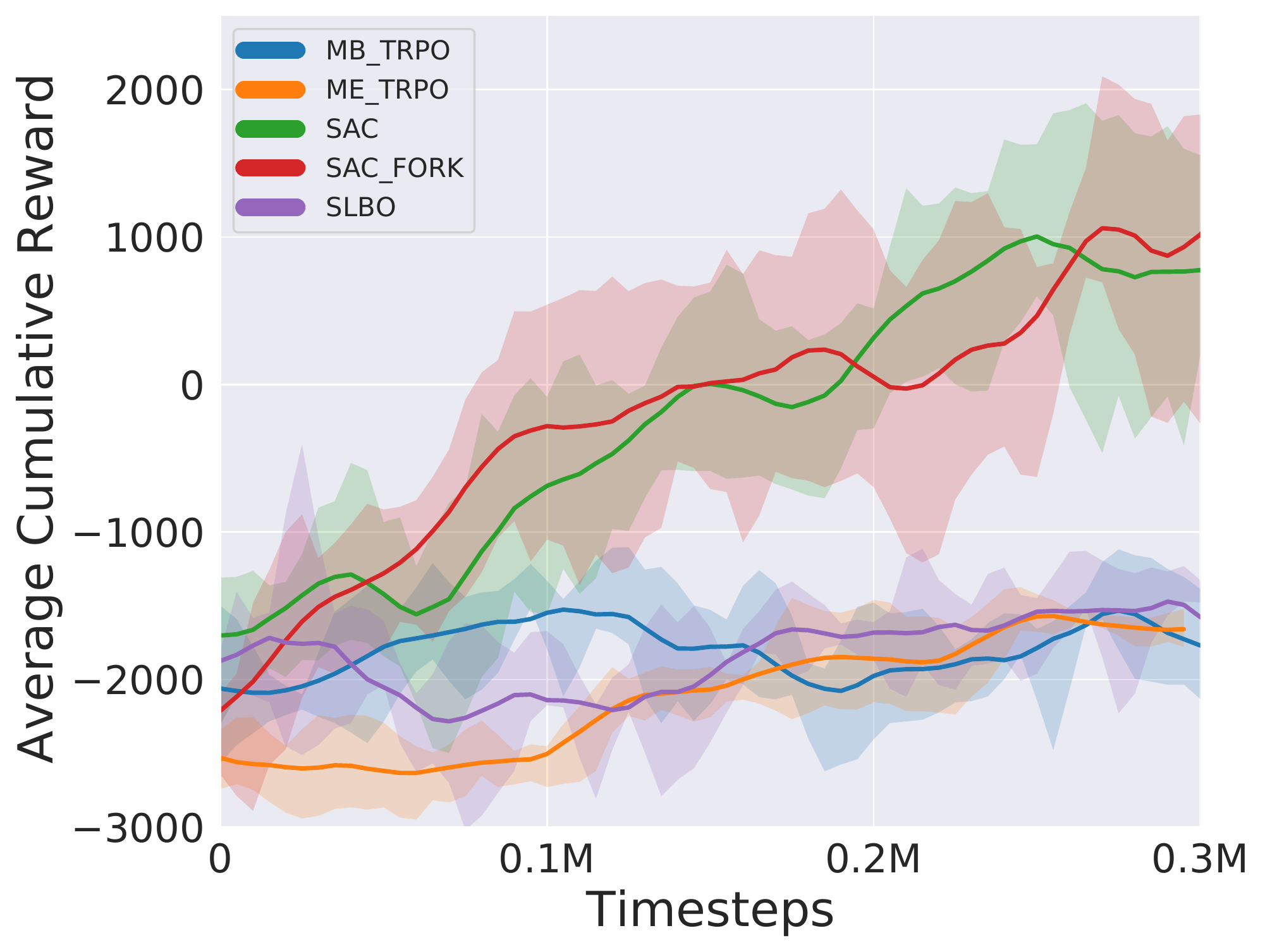}}}\
	\subfloat[HalfCheetah]{{\includegraphics[scale=0.35]{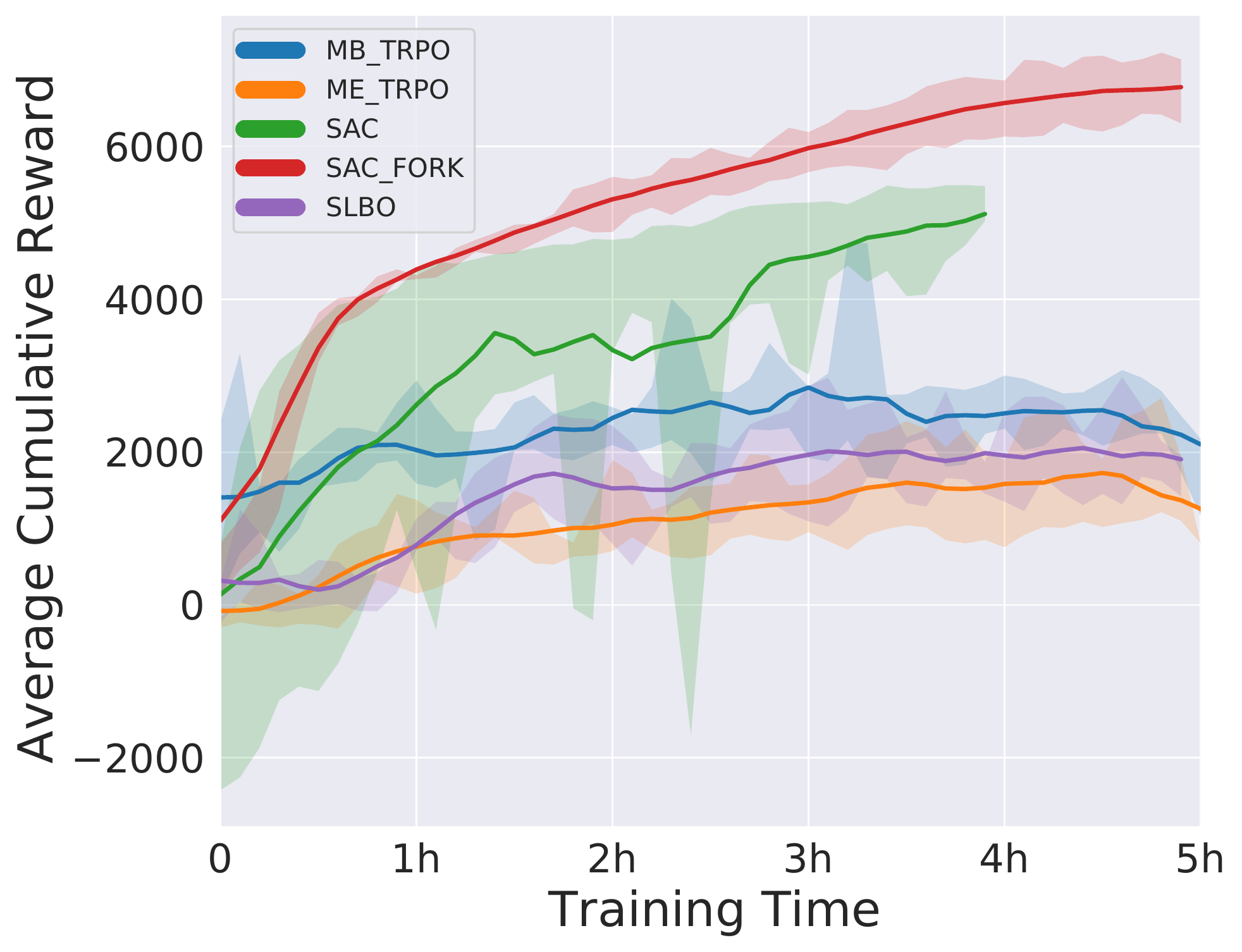} }}
	\subfloat[Walker2d]{{\includegraphics[scale=0.35]{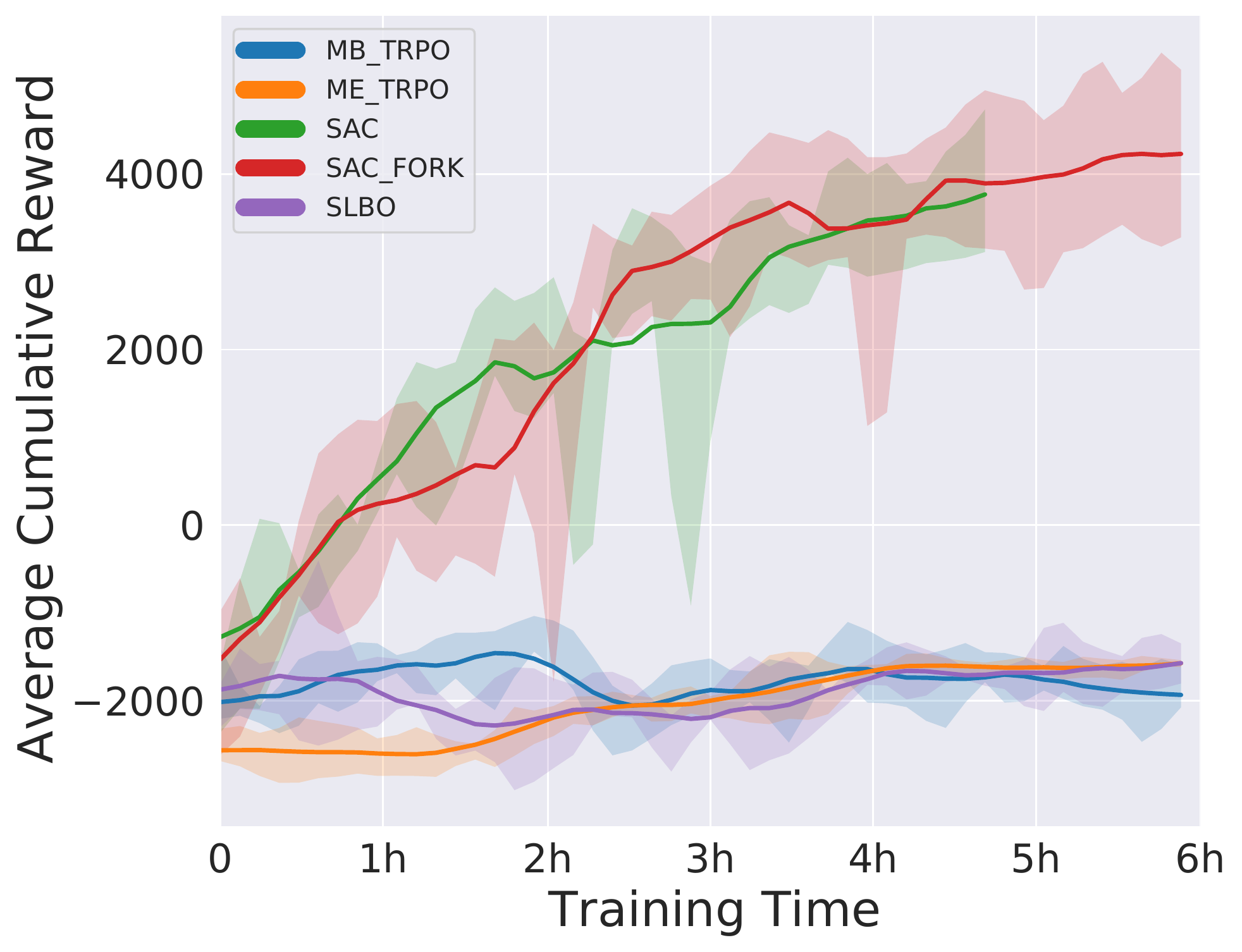} }}

	\caption{Learning curves of model-based algorithms versus FORK}
	\label{fig:model-based}
	\vspace{-0.2in}
\end{figure}

Figure \ref{fig:model-based}(a,b) show that the performance of SAC-FORK outperforms other model-based algorithms when training over the same number of timesteps. Note that FORK does not require extra training samples except those collected when interacting with environment, while other model-based RL algorithms have to be trained with imagined data at each step. Therefore, the training time of FORK is much shorter than model-based algorithms with the same number of training steps. For example, for Halfcheetah, SAC-FORK (SAC) only needs roughly 1.5 hours (1.2 hours) to finish 0.3M steps of training, but ME-TRPO needs {\bf 23 hours} and SLBO requires {\bf 9 hours}, respectively. The time complexity of FORK is shown in Figure \ref{fig:model-based}(c,d), where we compared the performance of the algorithms when being trained for the same amount of time. We can clearly observe that SAC-FORK significantly outperform model-based algorithms. 

\subsection{BipedalWalker-Hardcore-v3}

A variation of TD3-FORK can also solve  a well-known difficult environment, BipedalWalker-Hardcore-v3, in as few as four hours using a single GPU. From the best of our knowledge, the known algorithm needs to train for days on a 72 cpu AWS EC2 with 64 worker processes taking the raw frames as input (\url{https://github.com/dgriff777/a3c_continuous}). You can view the performance on BipedalWalkerHardcore-v3 during and after training at \url{https://youtu.be/pzzP8fA5Ipg}. The implementation details can be found in Appendix \ref{appsec:hardcore}. 

%% file: conclusion.tex
\section{Conclusions}
This paper proposes FORK, forward-looking Actor, as an add-on to Actor-Critic algorithms. The evaluation of six environments  demonstrated the significant performance improvements by adding FORK to two state-of-the-art model-free reinforcement learning algorithms. FORK also outperforms other model-based algorithms in both performance and time complexity. A variation of TD3-FORK further solved BipedalWalkerHardcore in as few as four hours with a single GPU. 

\section{Acknowledgements}
We gratefully thank Dr. Nikolay Gudkov at ETH Z{\"u}rich for the helpful comments. 

%% file: app_arxiv.tex
\onecolumn
\appendix
This appendix provides additional details about FORK and additional experiments. The appendix is organized as follows: 
\begin{itemize}
    \item In Section \ref{appsec:fork}, we provide additional details about FORK used in our experiments, including a description of TD3-FORK and the hyperparamters. 
    \item In Section \ref{appsec:more}, we present additional experimental results, including additional statistics of the experiments conducted in Section D, performance comparison under the same number of Actor+Critic training, and performance of different implementations of FORK.  
    \item In Section \ref{appsec:hardcore}, we present additional changes we made when using a variation of TD3-FORK, in particular, TD3-FORK-DQ, to solve the BipedalWalkerHardcore-v3. 
\end{itemize}

\section{Revised Reward Network}
\label{appsec:revised-reward}
\begin{figure}[htb]
    \centering
    \includegraphics[width=6cm,height=4cm]{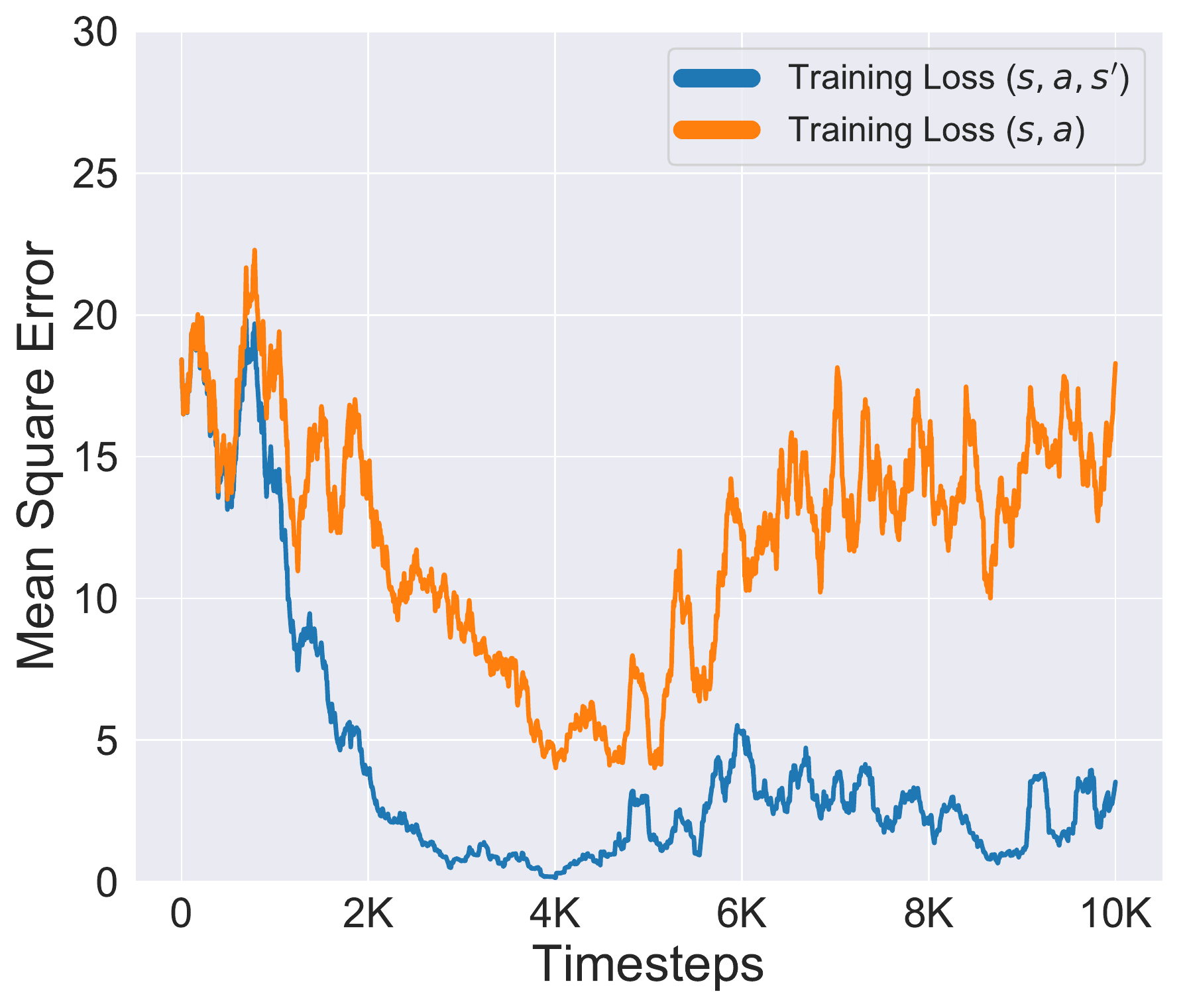}
    \caption{Training losses under the two different reward networks}
    \label{fig:r_loss}
\end{figure}

We found from our experiments that the reward network can more accurately predict reward $r_t$ when including the next state $s_{t+1}$ as input into the reward network.  Figure \ref{fig:r_loss} shows the mean-square-errors (MSE) of the reward network with $(s_t,a_t)$ as the input versus with  $(s_t, a_t, s_{t+1})$ as the input for BipedalWalker-v3 during the first 10,000 steps. We can clearly see that MSE is lower in the revised reward network. 

\section{Adaptive Weights versus Fixed Weights}
\label{appsec:fixed-adaptive}
We compared TD3-FORK with the fixed weights, named as TD3-FORK-F, where the weight is chosen to be 0.4. TD3-FORK performs the best in four out of the six environments. TD3-FORK-F has a worse performance than TD3 on Walker2d-v3. We therefore proposed and used the adaptive weight because of this observation. 

\begin{figure}[htb]
    \centering
    \subfloat{{\includegraphics[scale=0.5]{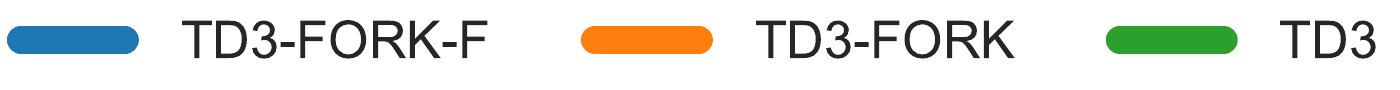} }}\quad
    \subfloat[BipedalWalker-v3]{{\includegraphics[scale=0.25]{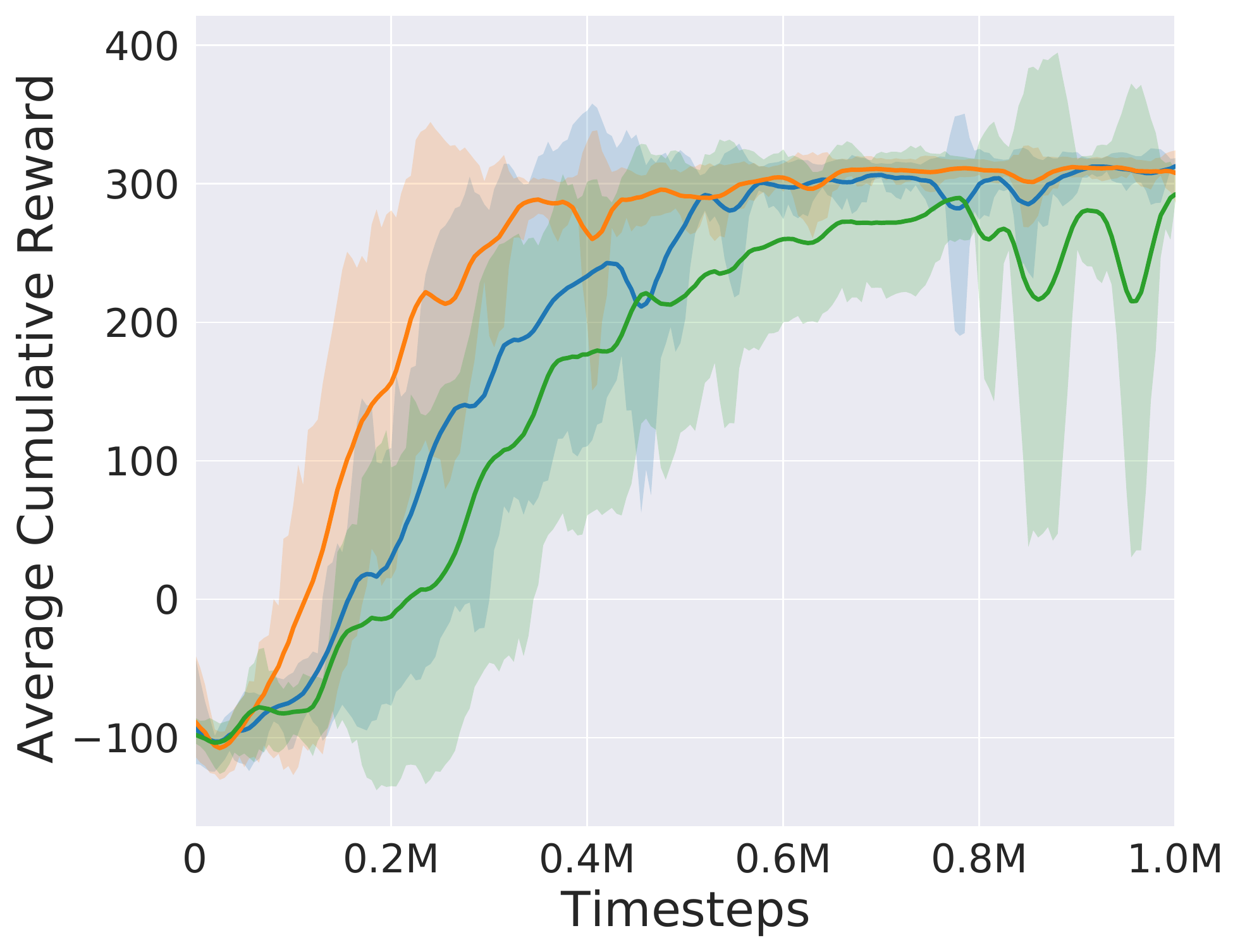} }}
    \subfloat[Ant-v3]{{\includegraphics[scale=0.25]{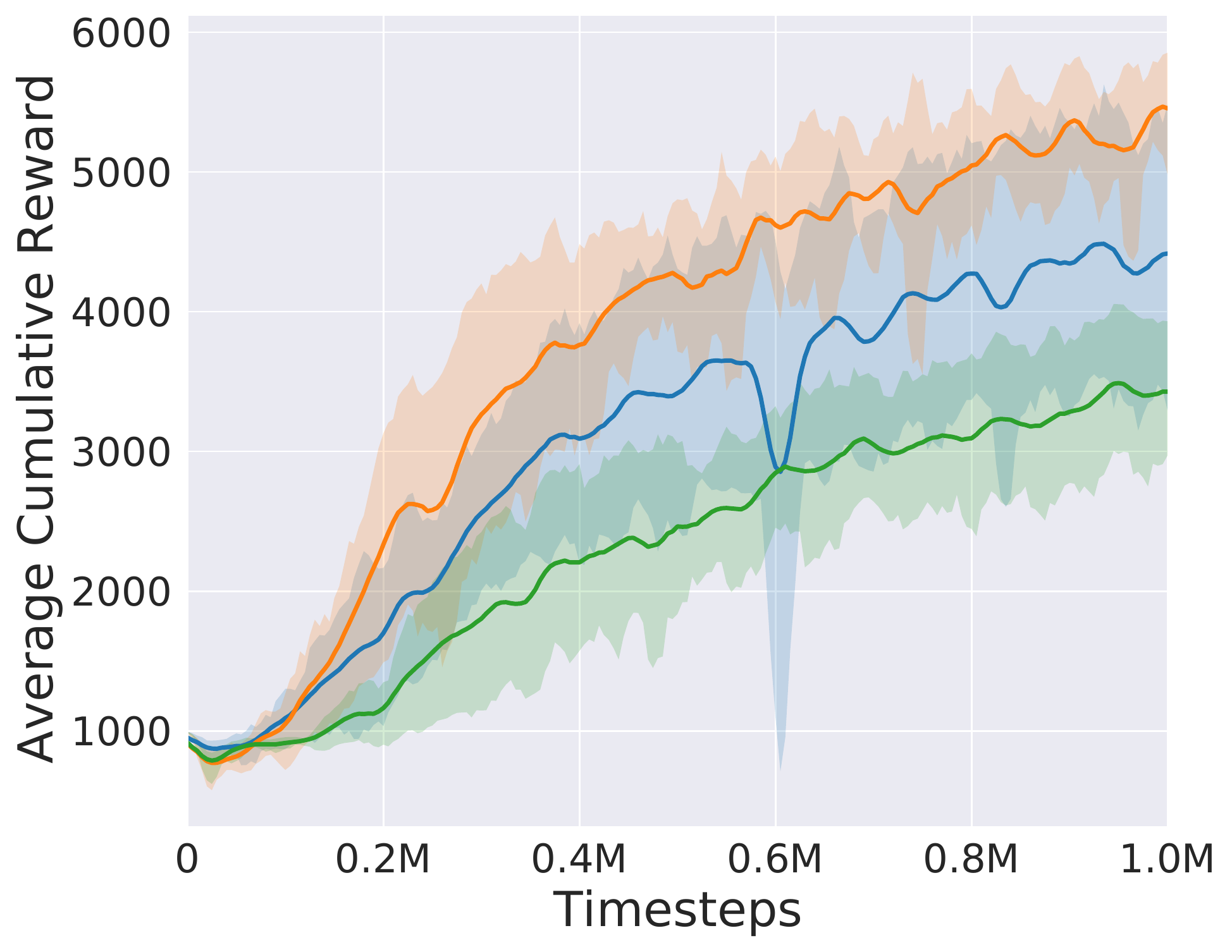} }}
    \subfloat[Hopper-v3]{{\includegraphics[scale=0.25]{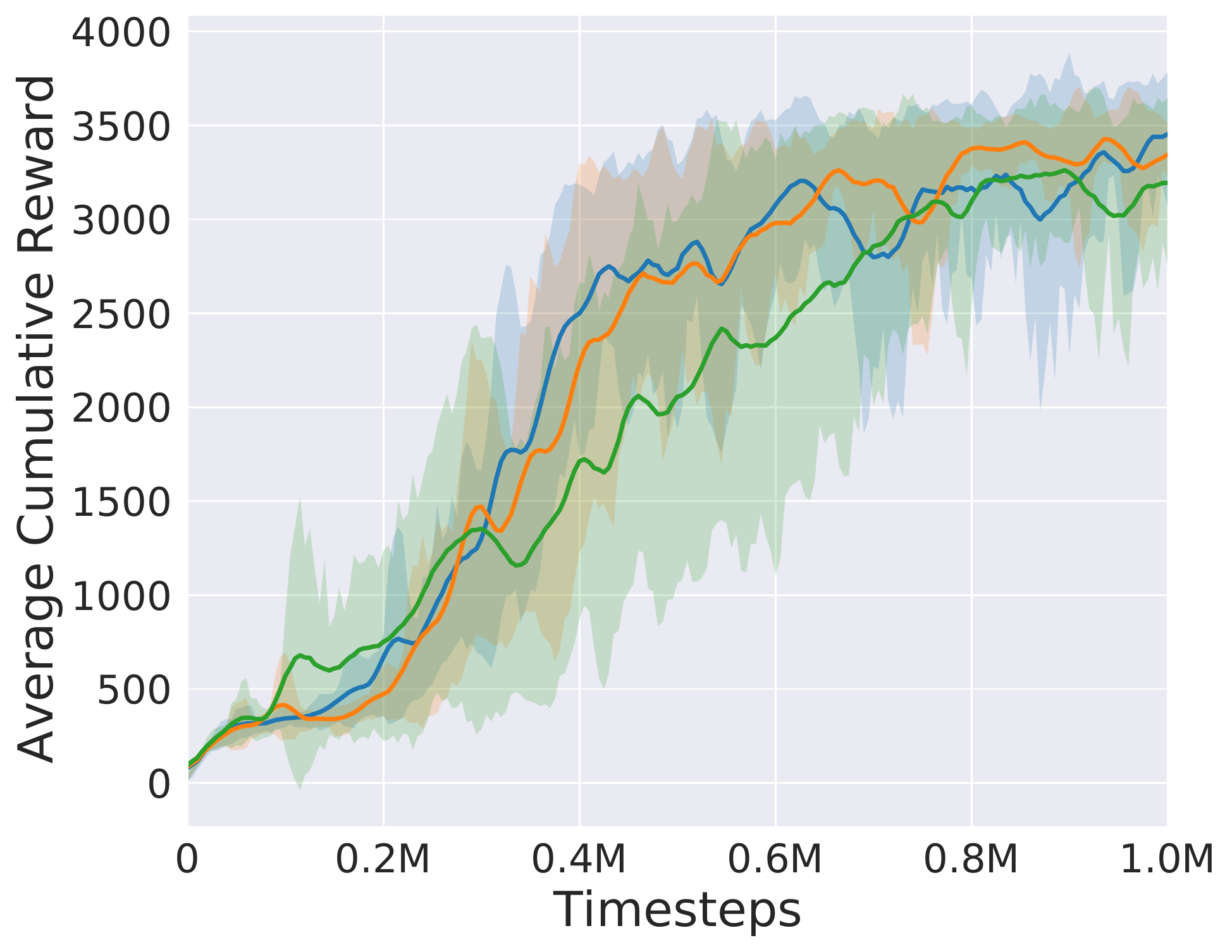} }}\\
    \subfloat[HalfCheetah-v3]{{\includegraphics[scale=0.25]{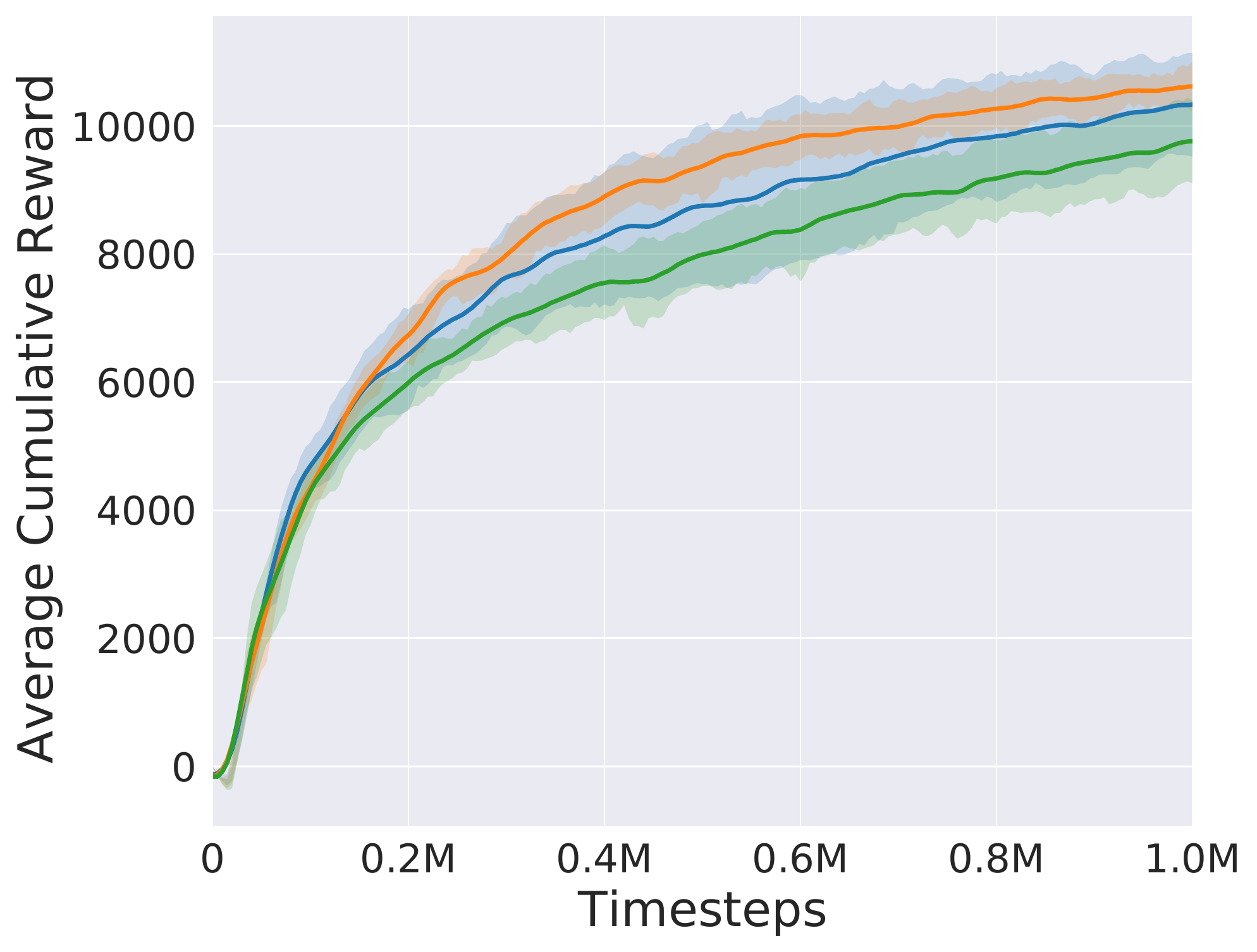}}}
     \subfloat[Humanoid-v3]{{\includegraphics[scale=0.25]{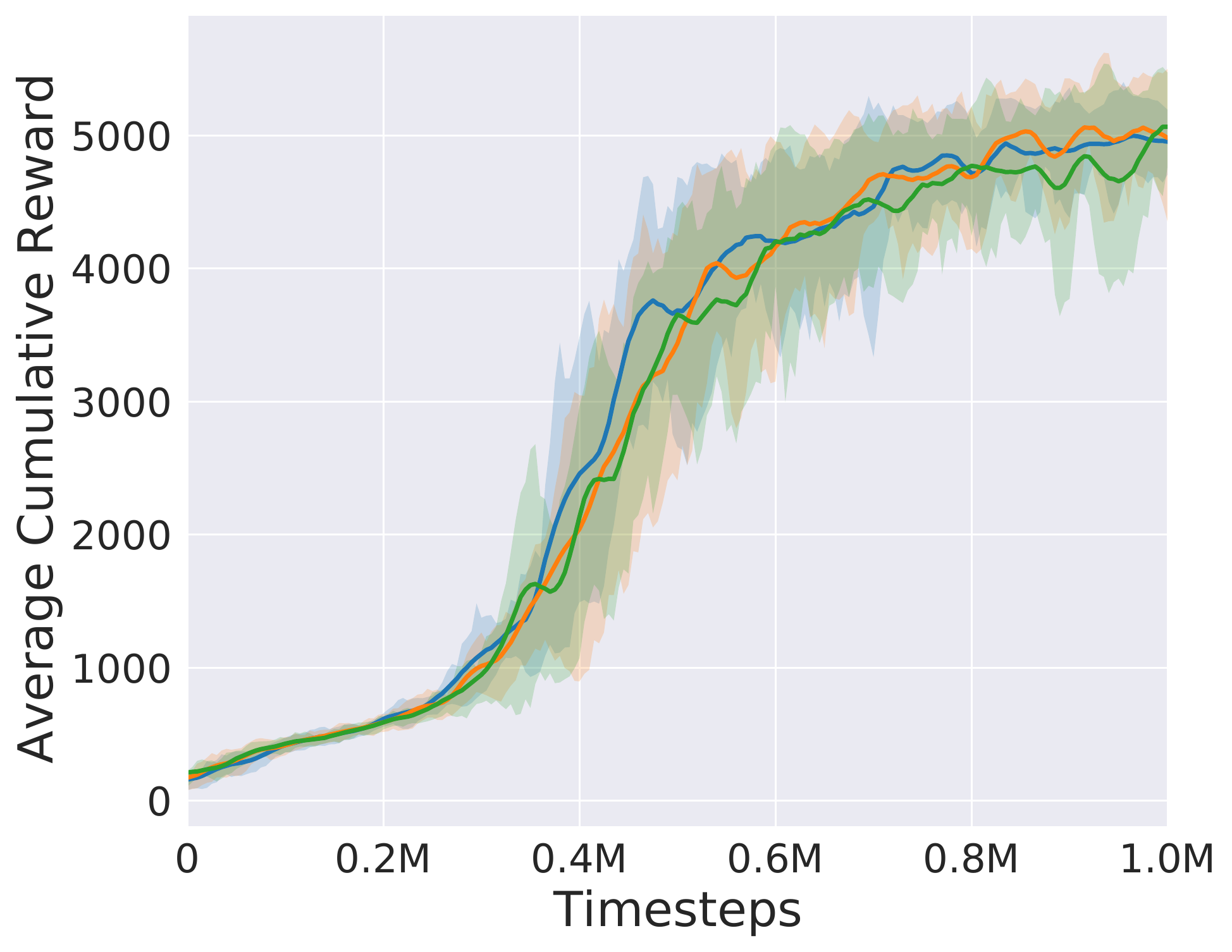} }}
    \subfloat[Walker2d-v3]{{\includegraphics[scale=0.25]{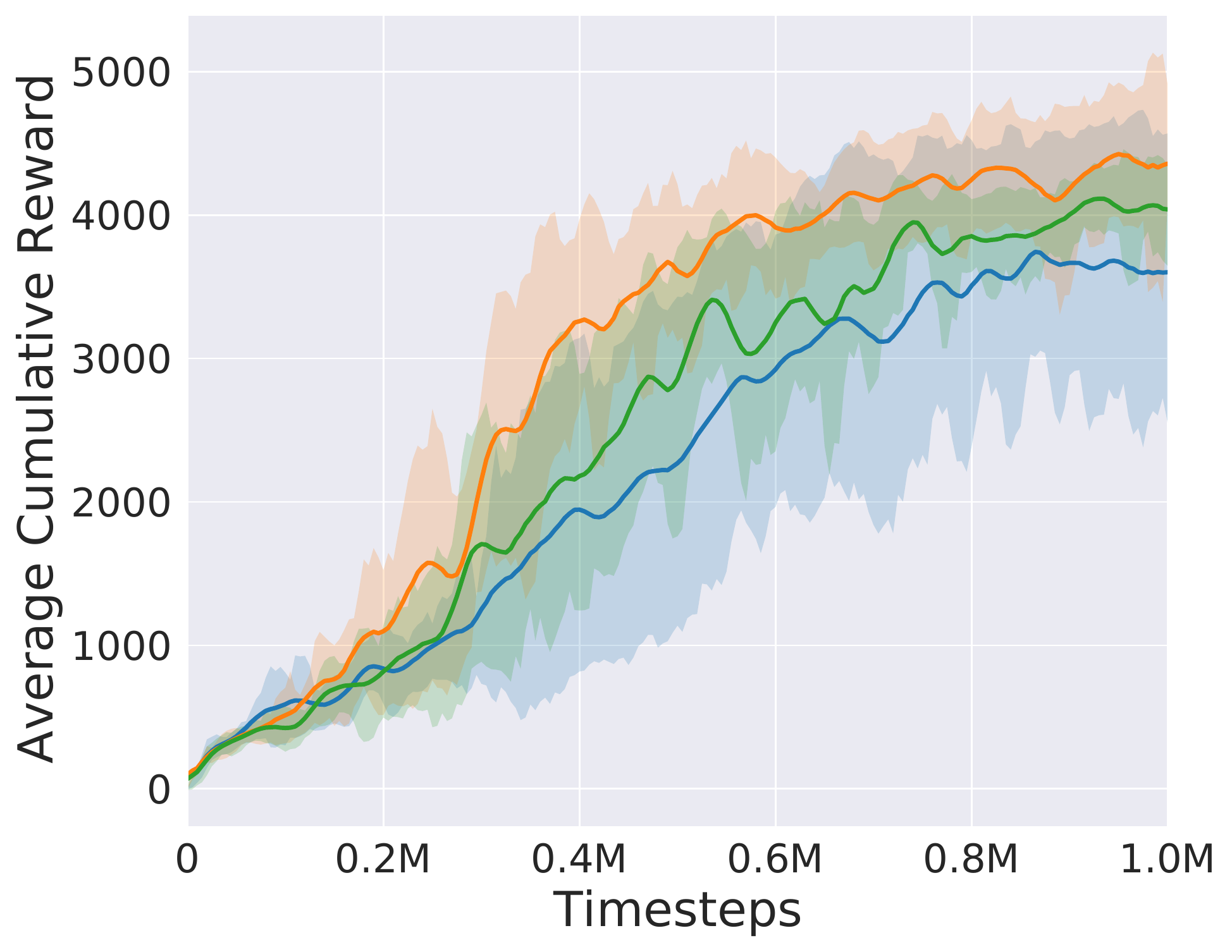} }}
    \caption{Learning curves of TD3, TD3-FORK-F and TD3-FORK. Curves are smoothed uniformly for visual clarity.}
    \label{fig:mujoco-td3-weight}
\end{figure}

\section{TD3-FORK}\label{appsec:fork}
\label{appsec:td3fork}
The detailed description of TD3-FORK can be found in Algorithm \ref{alg:td3-fork}.

  \begin{algorithm}[ht]
  \caption{TD3-FORK}\label{alg:sys_q}
  \begin{algorithmic}[1]
  \Statex Initialize critic networks $Q_{\psi_1},Q_{\psi_2}$ system network $F_{\theta},R_{\eta}$ and actor network $A_{\phi}$ with random parameters $\psi_1,\psi_2,\theta,\eta,\phi$
    \State Initialize target networks $\phi^\prime\leftarrow\phi,\psi_1^\prime\leftarrow\psi_1,\psi_2^\prime\leftarrow\psi_2$
  \Statex Initialize replay buffer $\mathcal{B},$ soft update parameter $\tau$
  \Statex Initialize base reward $r_0,w_0,$ threshold $\bar{l}$ and moving average reward $\bar{r}\leftarrow 0$
  \Statex Initialize noise clip bound $c,$ state bound $(o_{\min},o_{\max})$
    \For{episode $e=1,\dots,M$}
    \State Initialize observation state $s_0$
    \State Initialize episode reward $r=0$
    \For{$t=1,\dots,T$}
        \State Select action $a_t$ according to the current policy and exploration noise $a_t\sim A_{\phi}(s)+\epsilon_t,$ where $\epsilon_t\sim \mathcal{N}(0,\sigma)$ 
        \State Execute action $a_t$ and observe reward $r_t,$ new state $s_{t+1}$ 
        \State Store transition tuple $(s_t,a_t,r_t,s_{t+1})$ into replay buffer $\mathcal{B}$
        \State Sample a random minibatch of $N$ transitions $(s_i,a_i,r_i,s_{i+1})$ from $\mathcal{B}$
        \State $\tilde{a}_i\leftarrow \pi_{\phi^\prime}(s_{i+1})+\epsilon,\epsilon\sim{\textit clip}(\mathcal{N}(0,\tilde{\sigma}),-c,c) $
        \State Set $y_i=r_i+\gamma \min_{j=1,2}Q_{\psi_i^\prime}(s_{i+1}) $
        \State $r\leftarrow r+r_t$
        \State Update critic network by minimizing the loss: $L(\psi)=\frac{1}{N}\sum_{j=1}^2\sum_i\left(y_i - Q_{\psi_j}(s_i,a_i)\right)^2$
        \State Update state system network by minimizing loss: $L(\theta)=\| s_{i+1}-F_\theta(s_i,a_i)\|_{\hbox{\scriptsize smooth L1}}$
        \State Update reward system network by minimizing the loss: $L(\eta)=\frac{1}{N}\sum_i\left(r_{i}-R_{\eta}(s_i,a_i,s_{i+1}) \right)^2$
        \If{$t$ mod $d$=0}
            \State Update $\phi$ by the sampled policy gradient:
            \If{$L(\theta) > \bar{l}$} 
            \State $\nabla_{\phi}L(\phi)=\frac{1}{N}\sum_i \nabla_{a}Q_{\psi_1}(s_i,a )|_{a=A_\phi(s_i)}\nabla_\phi A_\phi(s_i) $
            \Else
            \State $s_{i+1}^\prime = {\textit clip}(F_\theta(s_i,A_\phi(s_{i})),o_{\min},o_{\max}),s_{i+2}^\prime = {\textit clip}(F_\theta(s_{i+1}^\prime,A_\phi(s_{i+1}^\prime)),o_{\min},o_{\max})$
            \State $\nabla_{\phi}L(\phi)=\frac{1}{N}\sum_i \left( \nabla_{a}Q_{\psi_1}(s_i,a)|_{a=A_\phi(s_i)}\nabla_\phi A_\phi(s_i) + w\nabla_a R_\eta(s_i,a,s_{i+1}^\prime)|_{a=A_\phi(s_i)}\nabla_\phi A_\phi(s_i)  \right.$
               \State $ \left.+w\gamma\nabla_a R_\eta(s_{i+1}^\prime,a,s_{i+2}^\prime)|_{a=A_\phi(s_{i+1}^\prime)}\nabla_\phi A_\phi(s_{i+1}^\prime)  +w\gamma^2 \nabla_{a}Q_{\psi_1}(s_{i+2}^\prime,a)|_{a=A_\phi(s_{i+2}^\prime)}\nabla_\phi A_\phi(s_{i+2}^\prime)   \right)$
            \EndIf
            \State Update target networks:
            \State $\phi^\prime\leftarrow \tau\phi + (1-\tau)\phi^\prime$
            \State $\psi_i^\prime\leftarrow \tau\psi_i +(1-\tau)\psi_i^\prime$
        \EndIf
    \EndFor
    \State Update $\bar{r} \leftarrow ((e-1)\bar{r} + r)/e$
    \State Update adaptive weight $w \leftarrow \min(1-\max(0,\frac{\bar{r}}{r_0}),1)w_0$
  \EndFor
  \end{algorithmic}
  \label{alg:td3-fork}
\end{algorithm}

\subsection{Hyperparameters}
\label{appsec:hyper}

Table \ref{tab:hyper} lists the hyper-parameter used in DDPG, SAC, SAC-FORK and TD3-FORK. We kept the same hyperparamter values used in SAC and TD3 codes provided or recommended by the authors. We did not tune these parameters because the goal is to show that FORK is a simple yet powerful add-on to existing Actor-Critic algorithms.

\begin{table}[htb]
    \centering
    \caption{Hyperparameters}
    \begin{tabular}{l|c}
       Parameter &  Value\\
       \hline
     Shared & \\
      \quad optimizer & Adam \\
      \quad  learning rate & $3\times 10^{-4}$\\
      \quad  discount $(\gamma)$ & 0.99 \\
      \quad replay buffer size & $10^6$ \\
      \quad number of hidden layers (all networks) & 2 \\
      \quad Batch Size & 100 \\
      \quad Target Update Rate & $5\times 10^{-3}$ \\
      \quad Target update delay (TD3, TD3-FORK) & 2\\
      \quad nonlinearity & ReLU \\
      \quad number of hidden units per layer (Critic) & 256 \\
      \quad number of hidden units per layer (Actor) & 256 \\
       \hline
      TD3-FORK & \\
      \quad number of hidden units of the system network & $[400,300]$ (Humanoid$[1024,1024]$) \\
      \quad number of hidden units of the reward network & $[256,256]$ (Humanoid$[1024,1024]$) \\
      \hline
      SAC-FORK & \\
      \quad number of hidden units of the system network & $[512,512]$ (Humanoid$[1024,1024]$) \\
      \quad number of hidden units of the reward network & $[512,512]$ (Humanoid$[1024,1024]$) \\
      \hline
    \end{tabular}
    \label{tab:hyper}
\end{table}

Table \ref{tab:hyper-2} summarizes the environment specific parameters. In particular, the base weight and base cumulative reward used in implementing the adaptive weight, and threshold for adding FORK. The base cumulative rewards for TD3-FORK are the typical cumulative rewards under TD3 after training 1 million steps. The base cumulative rewards for SAC-FORK are similarly chosen but with a more careful tuning. The thresholds are the typical loss values after training the system networks for about 20,000 times including the first 10,000 exploration steps. In the implementation, FORK is added to Actor training only after the system network can predict the next state reasonably well. We observed TD3-FORK with our intuitive choices of hyperparameters worked well across different environments and required little tuning, while SAC-FORK required some careful tuning on choosing the base weights and the base cumulative rewards.

\begin{table}[htb]
    \centering
    \caption{Environment Specific Parameters}
    \begin{tabular}{lcccc}
    \hline
     Environment & Base Weight $w_0$ &  Base Cumulative Reward $r_0$ & System threshold $\bar{l}$ \\ 
       \hline
      TD3-FORK\\
    \quad BipedalWalker-v3 & $0.6$& $320$& $0.01$ \\
    \quad Ant-v3 & $0.6$& $6200$& $0.15$ \\
    \quad Hopper-v3 &$0.6$& $3800$& $0.0020$ \\
    \quad HalfCheetah-v3 &$0.6$& $12000$& $0.20$ \\
    \quad Humanoid-v3 &  $0.6$ & $5200$& $0.20$ \\
    \quad Walker2d-v3& $0.6$& $4500$& $0.15$ \\
      \hline
     SAC-FORK\\
    \quad BipedalWalker-v3 & $0.40$& $320$& $0.01$ \\
    \quad Ant-v3 & $0.40$& $5200$& $0.020$ \\
    \quad Hopper-v3 &$0.40$& $4000$& $0.0020$ \\
    \quad HalfCheetah-v3 &$0.10$& $8000$& $0.10$ \\
    \quad Humanoid-v3 &  $0.10$ & $4500$& $0.10$ \\
    \quad Walker2d-v3& $0.30$& $3500$& $0.15$ \\
      \hline
    \end{tabular}
    \label{tab:hyper-2}
\end{table}

\section{Additional Experimental Results} \label{appsec:more}
\subsection{Best Average Cumulative Reward, Standard-Deviation, and Best Instance Cumulative Reward}
Table \ref{tab:max_average} summarizes the best average cumulative rewards, the associated standard-deviations, and best instance cumulative rewards. They are defined as follows. Recall that each algorithm is trained for five instances, where each instance includes 1 million steps of training. During the training process, we evaluated the algorithm every 5,000 steps without the exploration noise. For each evaluation, we calculated the average cumulative rewards (without discount)  over 10 episodes, where each episode is $0\sim1,600$ under BipedalWalker-v3, is $0\sim 1,000$ under Ant-v3, Walker2d-v3, Hopper-v3, Humanoid-v3, and is exactly 1,000 under HalfCheetah-v3. 

Now let $X^{(l)}_\tau$ denote the average cumulative reward at the $\tau$th evaluation during the $l$th instance. Then 
\begin{align*}
    \hbox{Best Average Cumulative Reward (Best Average): } &\max_\tau \frac{1}{5}\sum_{l=1}^5 X^{(l)}_\tau\\
    \hbox{Standard-Deviation: } & \sqrt{\frac{1}{5}\sum_{l=1}^5\left(X^{(l)}_\tau-\overline{X}_\tau \right)^2 } \\
    \hbox{Best Instance Cumulative Reward (Best Instance): }  &\max_l \max_\tau X^{(l)}_\tau\\
\end{align*}

\onecolumn
\begin{table}[htb]
    \centering
    \caption{Best Average Cumulative Rewards, Standard Deviations, and Best Instance Cumulative Rewards of TD3-FORK, TD3, DDPG, SAC, SAC-FORK over Six Environments. The Best Value for Each Environment is in Bold Text.}
    \begin{tabular}{lccccc}
     \hline
        {\bf Environment} &  {\bf TD3-FORK} & {\bf TD3} & {\bf DDPG} & {\bf SAC} & {\bf SAC-FORK} \\
    \hline
    BipedalWalker-v3 \\
    Best Average Cumulative Reward & ${ 317.73}$   & $307.97$ &   $139.29$  & $313.49$ & ${\bf 319.47}$\\
    Standard Deviation & ${\pm 4.78}$   & $\pm 11.28$ &   $\pm 125.49$  & $\pm 9.03$ & $\pm 6.35$\\
    Best Instance Cumulative Reward& ${323.57}$   & $317.47$ &   $260.51$  & $322.40$ & ${\bf 325.13}$\\
    \hline
    Ant-v3 \\
    Best Average Cumulative Reward& ${\bf 5731.85}$ & $3678.75$   &  $995.59$  &  $3600.24$ &  $5033.09$  \\
    Standard Deviation& ${ \pm 229.64}$ & $\pm 485.37$   &  $\pm 2.67$  &  $\pm 847.70$ &  $\pm 628.85$  \\
    Best Instance Cumulative Reward&${\bf 6035.45}$ & $4546.48$   &  $999.79$  &  $4634.73$ &  $5634.70$  \\
    \hline
    Hopper-v3 \\
    Best Average Cumulative Reward & ${ 3539.95} $ & $3526.67 $ & $1738.05$ &  $3504.41$ & ${\bf 3568.20}$ \\
    Standard Deviation & ${\pm 119.34} $ & $\pm 69.62 $ & $\pm 944.10$ &  $\pm 159.02$ & $\pm 97.83$ \\
    Best Instance Cumulative Reward& ${ 3691.80} $ & $3591.42$ & $3573.53$ &  $3674.43$ & ${\bf 3700.20}$ \\
    \hline
    HalfCheetah-v3 \\
    Best Average Cumulative Reward & ${10875.32} $   & $9899.12$  & $5975.80 $ & $11230.00$ & ${\bf 11604.60}$\\
    Standard Deviation & $ \pm 205.55 $   & $\pm 669.13$  & $\pm 2500.40 $ & $\pm 220.06$ & $\pm 217.55$\\
    Best Instance Cumulative Reward & ${ 11172.02} $   & $10361.92$  & $9722.45 $ & $11501.31$ & ${\bf 11738.65}$\\
    \hline
     Humanoid-v3\\
     Best Average Cumulative Reward & ${5439.31} $  & $5394.02$    & $465.10$ & $5484.04 $ & ${\bf 5509.89}$ \\
     Standard Deviation & ${\pm 152.26} $  & $\pm 117.84$    & $\pm 78.26$ & $\pm 72.94 $ & $\pm 103.72$ \\
    Best Instance Cumulative Reward& ${\bf 5685.77} $  & $5534.09$    & $529.84$ & $5616.20$ & ${ 5684.92}$ \\
    \hline
    Walker2d-v3\\
    Best Average Cumulative Reward &  ${ 4619.64 }$   &    $4319.09 $ &  $ 2505.23  $ & $4763.36 $ & ${\bf 5093.01}$  \\
     Standard Deviation &  ${ \pm 497.84}$   &    $\pm 218.34 $ &  $ \pm 1354.82   $ & $\pm 712.07 $ & $ \pm  444.38$  \\
   Best Instance Cumulative Reward &  ${ 5198.83}$ &  $4574.02$ &  $4618.44 $ & $5937.50 $ & ${\bf 5945.80}$ \\
    \hline
    \end{tabular}
    \label{tab:max_average}
\end{table}

\subsection{Different Implementations of FORK}
\label{appsec:forks}
As we mentioned in Section \ref{sec:alg}, FORK can have different implementations. We considered two examples in Section \ref{sec:eval}, and compared their performance as add-on to TD3. We call FORK with loss function \eqref{loss:fork-s} FORK-S, standing for single-step FORK; call FORK with loss function \eqref{loss:fork-dq} and $w^\prime=0.5$ FORK-DQ, standing for Double-Q FORK; and  FORK with loss function \eqref{loss:fork-dq} and $w^\prime=0$ FORK-Q,  standing for Q FORK. From Table \ref{tab:max_forks}, we can see that in terms of best average cumulative reward, TD3-FORK performs the best four out of the six environments and TD3-FORK-S performs the best in the remaining two. This is the reason we selected the current form of FORK. 

\begin{figure}[htb]
    \centering
    \subfloat{{\includegraphics[scale=0.5]{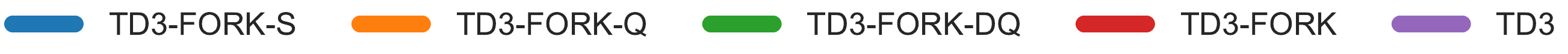} }}\quad
    \subfloat[BipedalWalker-v3]{{\includegraphics[scale=0.25]{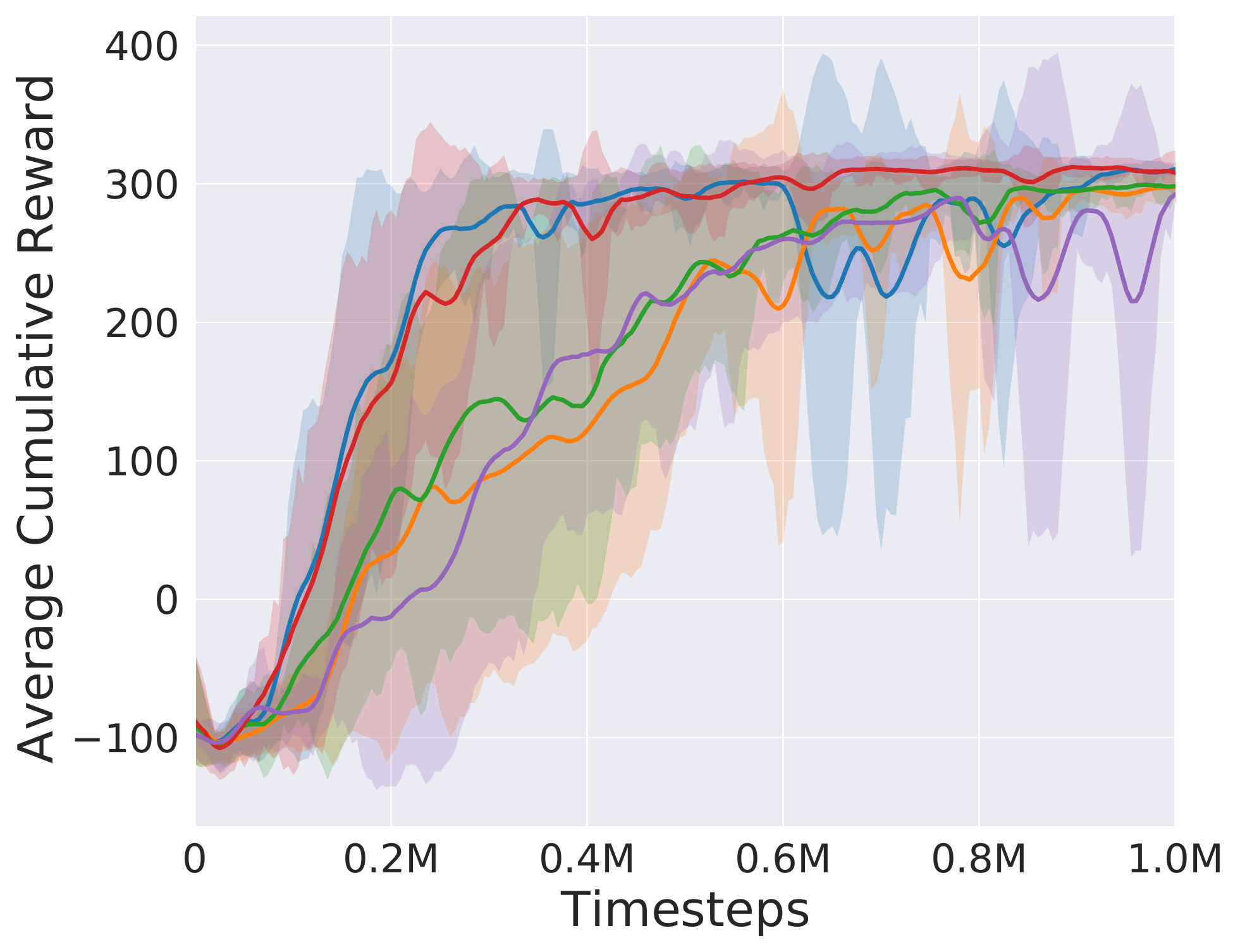} }}
    \subfloat[Ant-v3]{{\includegraphics[scale=0.25]{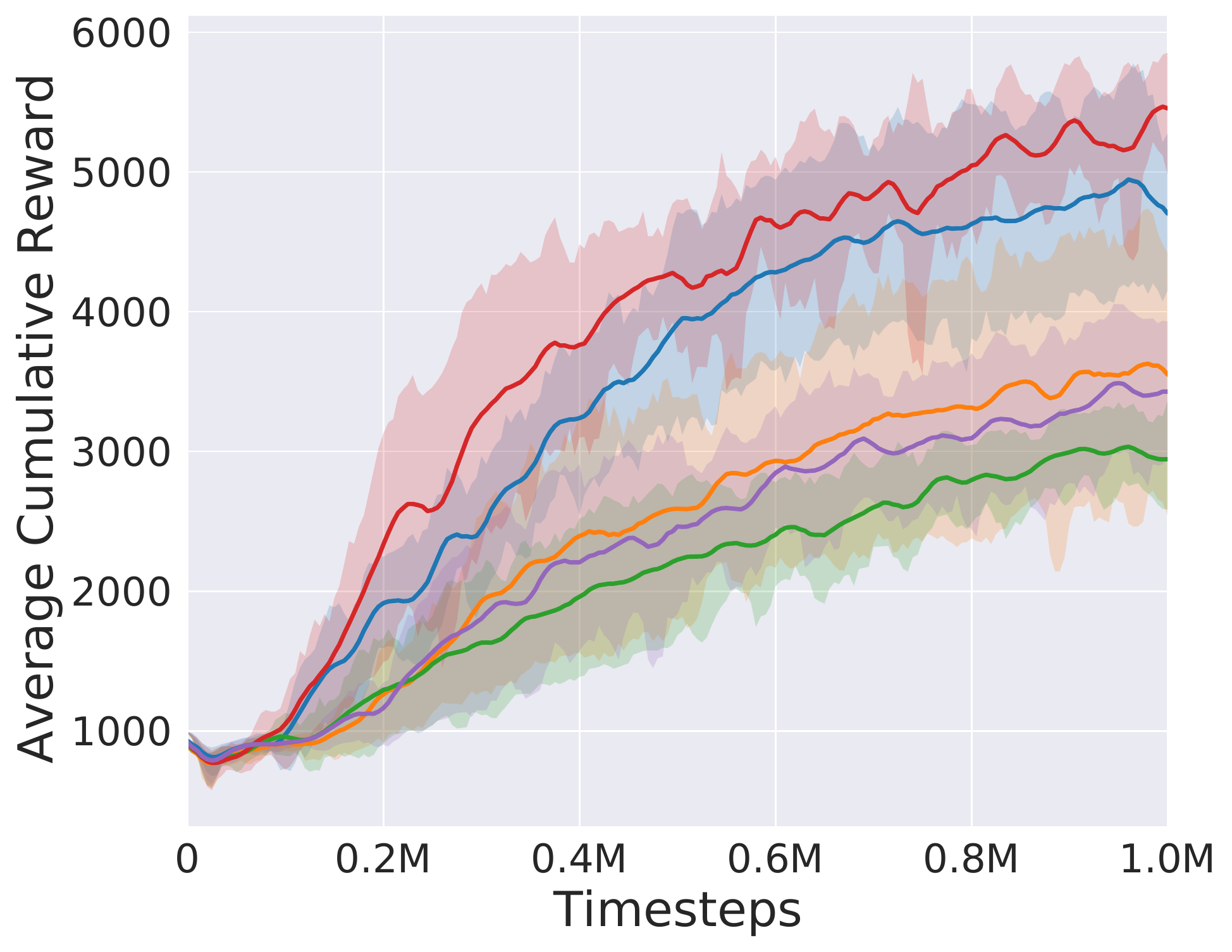} }}
    \subfloat[Hopper-v3]{{\includegraphics[scale=0.25]{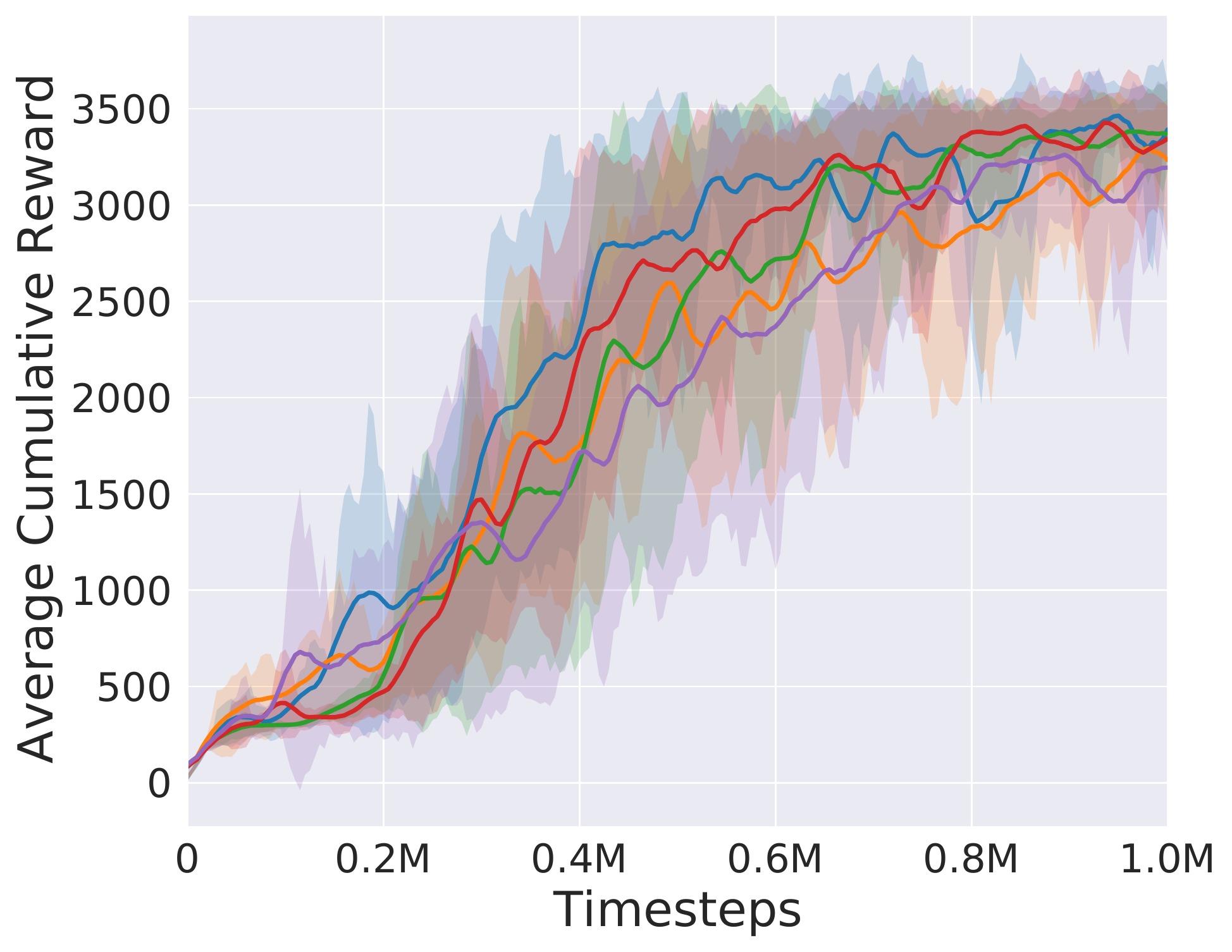} }}\\
    \subfloat[HalfCheetah-v3]{{\includegraphics[scale=0.25]{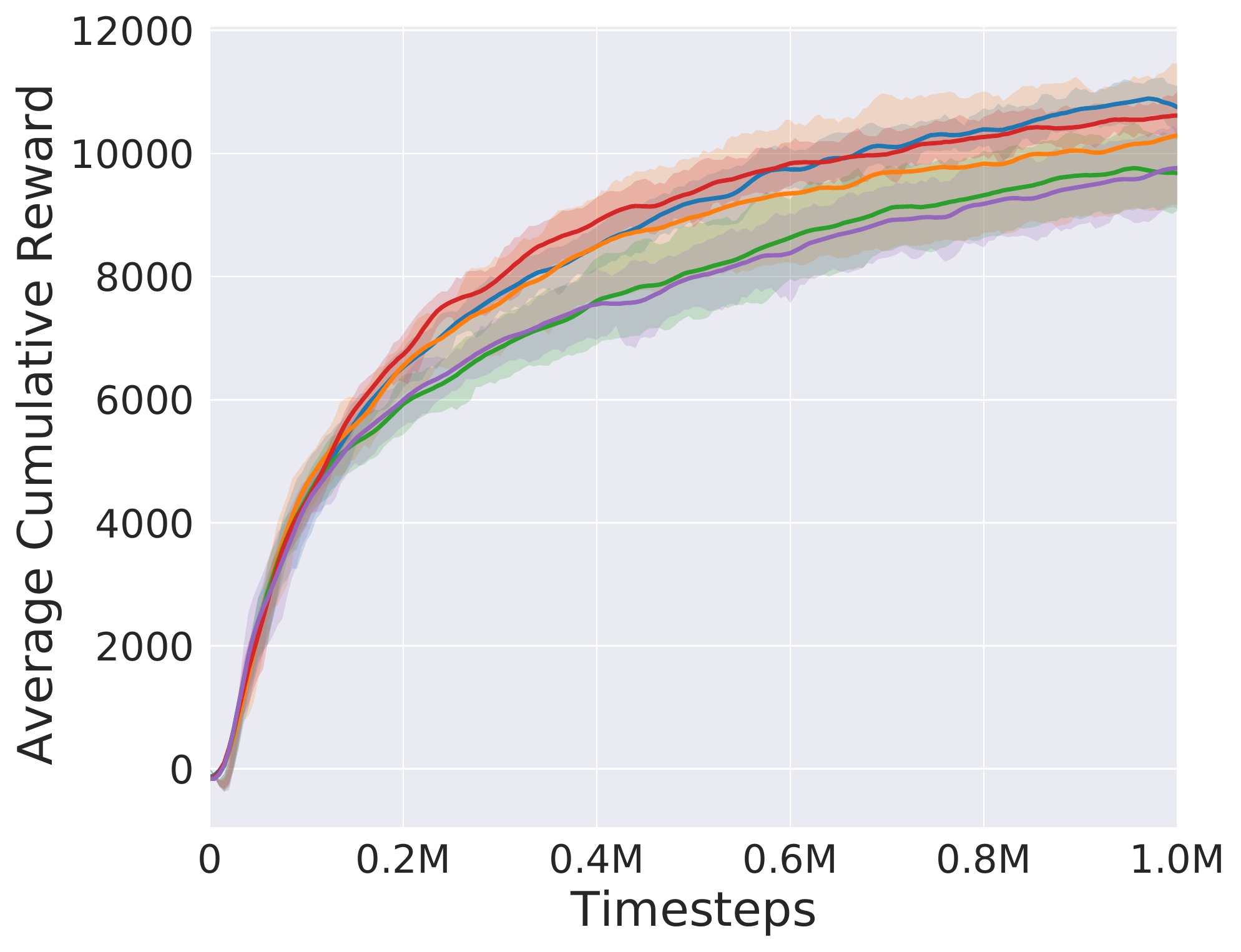}}}
     \subfloat[Humanoid-v3]{{\includegraphics[scale=0.25]{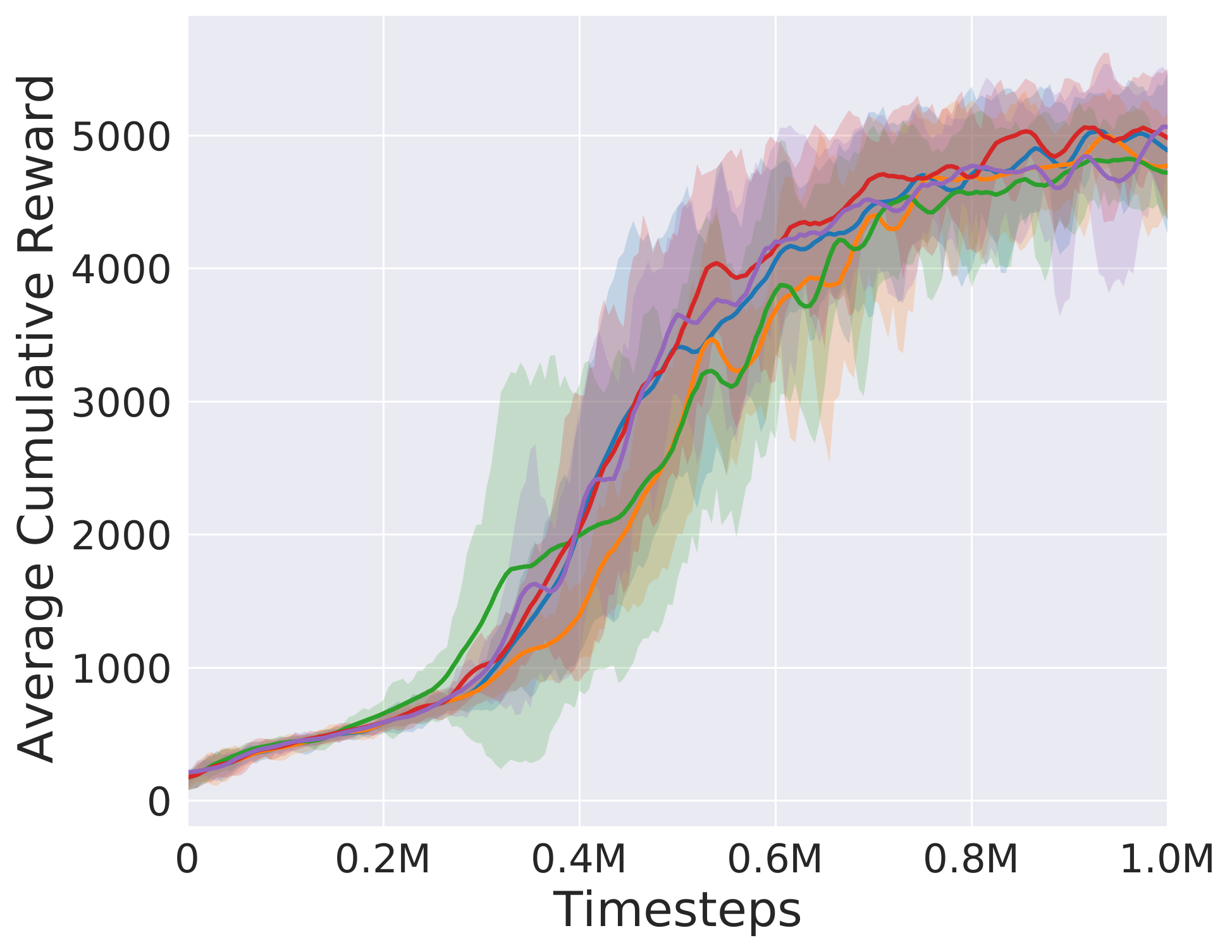} }}
    \subfloat[Walker2d-v3]{{\includegraphics[scale=0.25]{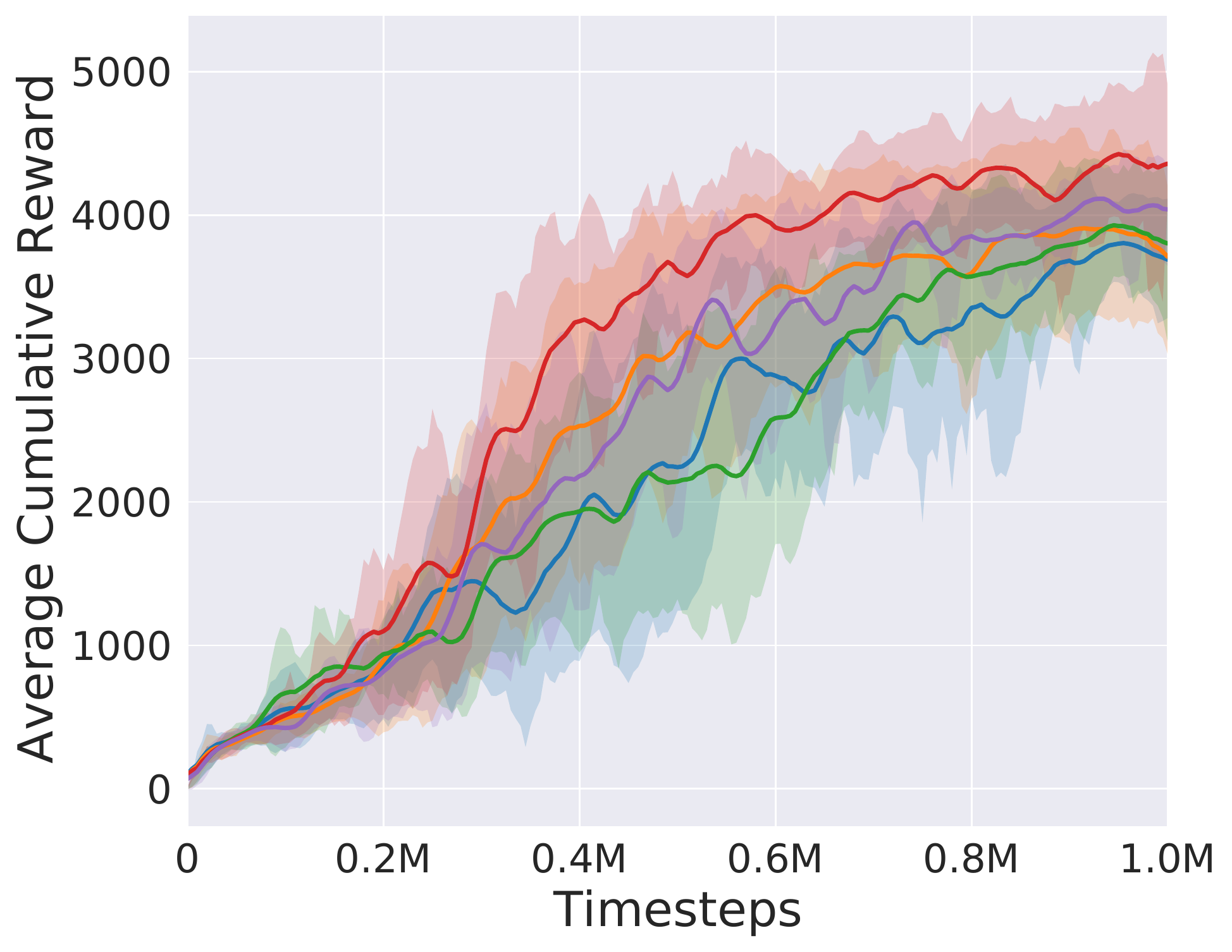} }}
    \caption{Learning curves of TD3-FORK, TD3, TD3-FORK-S, TD3-FORK-Q and TD3-FORK-DQ. Curves are smoothed uniformly for visual clarity.}
    \label{fig:mujoco-td3-forks}
\end{figure}

\begin{table}[ht]
    \centering
    \caption{Best Average Cumulative Rewards, Standard Deviations, and Best Instance Cumulative Rewards of TD3-FORK, TD3, TD3-FORK-S, TD3-FORK-Q, and TD3-FORK-DQ over Six Environments. The Best Values are in Bold Text.}
    \begin{tabular}{lccccc}
     \hline
        {\bf Environment} &  {\bf TD3-FORK} & {\bf TD3} & {\bf TD3-FORK-S} & {\bf TD3-FORK-Q} & {\bf TD3-FORK-DQ}\\
    \hline
    BipedalWalker-v3 \\
    Best Average & ${\bf 317.73}$   & $307.69$   & $314.63$ & $302.96$ & $306.44$ \\
    Standard Deviation & ${\pm 4.78}$   & $\pm 11.28$      & $\pm 3.61$ & $\pm 9.17$ & $\pm 5.98$  \\
    Best Instance& ${\bf 323.57}$   & $317.47$  & $320.11$ & $313.93$  & $ 315.55$\\
    \hline
    Ant-v3 \\
    Best Average& ${\bf 5731.85}$ & $3678.75$   &  $5226.37$ &  $ 3905.39$ &$3287.929$  \\
    Standard Deviation& ${ \pm 229.64}$ & $\pm 485.37$  &  $\pm  748.03$ &  $\pm  1019.78$  & $\pm 220.10$ \\
    Best Instance&${\bf 6035.45}$ & $4546.48$  &  $5748.94$ &  $ 5563.98$ & $3499.86$ \\
    \hline
    Hopper-v3 \\
    Best Average& ${ 3539.95} $ & $3526.67 $ &   ${\bf 3605.09}$ & $ 3426.54$ & $3482.14$ \\
    Standard Deviation & ${\pm 119.34} $ & $\pm 69.62$  &  $\pm 89.04$ & $\pm 119.51$  & $\pm 136.36$\\
    Best Instance& $3691.80 $ & $3591.42$  &  ${\bf 3740.71}$ & $3588.89$  & $3677.66$\\
    \hline
    HalfCheetah-v3 \\ 
    Best Average& $ 10875.32 $   & $9899.12$   & ${\bf 11077.10}$ & $10405.40$ & $9942.23$ \\
    Standard Deviation & $\pm 205.55 $   & $\pm 669.13$ & $\pm 337.05 $ & $\pm 1154.00$ & $\pm 675.52$ \\
    Best Instance& $ 11172.02 $   & $10361.92$  & $11450.83$ & ${\bf 11524.62}$ & $10668.92$\\
    \hline
     Humanoid-v3\\
     Best Average& $ {\bf 5439.31} $  & $5394.02$   & $ 5345.92 $ & $5255.54$ & $5270.73$ \\
     Standard Deviation & ${\pm 152.26} $  & $\pm 117.84$  & $\pm 204.92 $ & $\pm 210.65$ & $\pm 96.97$ \\
    Best Instance& ${5685.77} $  & $5534.09$ & ${\bf 5706.35}$ & $5669.17$ & $5499.36$ \\
    \hline
    Walker2d-v3\\
    Best Average&  ${\bf 4619.46 }$   &    $4319.09 $ &  $4089.02$ & $ 4175.40$  & $ 4177.34$ \\
     Standard Deviation &  ${ \pm 497.84}$   &    $\pm 218.34 $ & $\pm 260.61 $ & $ \pm 601.82$ & $\pm 372.90$  \\
   Best Instance&  ${\bf 5198.83}$ &  $4541.51$ &  $4474.02$ & $4934.03$ & $4920.74$ \\
    \hline
    \end{tabular}
    \label{tab:max_forks}
\end{table}

\section{BipedalWalkerHardcore}
\label{appsec:hardcore}
TD3-FORK-DQ can solve the difficult BipedalWalker-Hardcore-v3 environment with as few as four hours. The hardcore version is much more difficult than BipedalWalker. For example, a known algorithm needs to train for days on a 72 cpu AWS EC2 with 64 worker processes taking the raw frames as input (\url{https://github.com/dgriff777/a3c_continuous}). TD3-FORK-DQ, a variation of TD3-FORK, can solve the problem in as few as four hours by using default GPU setting provided by Google Colab\footnote{https://colab.research.google.com/notebooks/intro.ipynb} and with sensory data (not images). 
The performance on BipedalWalkerHardcore-v3 during and after training can be viewed at \url{https://youtu.be/pzzP8fA5Ipg}. 

To solve BipedalWalkerHardcore, we made several additional changes. 
\begin{itemize}
    \item[(i)] We changed the -100 reward to -5.
    \item[(ii)] We increased other rewards by a factor of 5.
    \item[(iii)] We implemented a replay buffer where failed episodes, in which the bipedalwalker fell down at the end, and successful episodes are added to the replay-buffer with 5:1 ratio.  
\end{itemize}

The changes to the rewards (i) and (ii) were suggested in the  blog \footnote{\url{https://mp.weixin.qq.com/s?__biz=MzA5MDMwMTIyNQ==&mid=2649294554&idx=1&sn=9f893801b8917575779430cae89829fb&scene=21\#wechat_redirect}}. Using reward scaling to improve performance has been also reported in \cite{henderson2017deep}. We made change (iii) because we found failed episodes are more useful for learning than successful ones. The reason we believe is that when the bipidedalwalker already knows how to handle a terrain, there is no need to further train using the same type of terrain. When the training is near the end, most of the episodes are successful so adding these successful episodes overwhelm the more useful episodes (failed ones), which slows down the learning.